\documentclass[]{fairmeta}

\title{Explore Theory-of-Mind: Program-Guided Adversarial Data Generation for Theory of Mind Reasoning}

\author[1,2,*]{Melanie Sclar}
\author[1]{Jane Yu}
\author[1]{Maryam Fazel-Zarandi}
\author[2]{Yulia Tsvetkov}
\author[1,3,*]{Yonatan Bisk}
\author[2]{Yejin Choi}
\author[1]{\ \ \ \ \ Asli Celikyilmaz}
\affiliation[1]{FAIR at Meta}
\affiliation[2]{University of Washington}
\affiliation[3]{Carnegie Mellon University}

\contribution[*]{Work done at Meta}

\usepackage{graphicx}
\usepackage{booktabs}
\usepackage{fvextra}
\usepackage{ulem}
\usepackage{multirow}
\usepackage{arydshln}
\usepackage{comment}

\usepackage{pifont}
\newcommand{\cmark}{\ding{51}}%
\newcommand{\xmark}{\ding{55}}%

\usepackage{amsmath,amsfonts,bm}

\def\eqref#1{equation~\ref{#1}}

\def\1{\bm{1}}

\DeclareMathAlphabet{\mathsfit}{\encodingdefault}{\sfdefault}{m}{sl}
\SetMathAlphabet{\mathsfit}{bold}{\encodingdefault}{\sfdefault}{bx}{n}

\usepackage{enumitem}

\usepackage{hyperref}
\usepackage{url}
\usepackage{comment}
\usepackage{ bbold }

\newcommand{\methodname}{\textsc{ExploreToM}}

\definecolor{easy}{RGB}{84, 174, 50}  
\definecolor{medium}{RGB}{255, 147, 0}
\definecolor{hard}{RGB}{238, 34, 12}
\definecolor{promptboxcolor}{RGB}{253, 240, 220} 
\definecolor{darkbrown}{rgb}{0.58, 0.42, 0.23}
\newcommand{\finalnumber}[1]{\textcolor{black}{#1}}

\newcommand{\myboxed}[1]{%
  \rlap{\hspace*{\dimexpr\fboxrule+\fboxsep\relax}%
    \phantom{\m@th$\displaystyle#1$}}%
    \smash{\boxed{#1}}}

\abstract{
Do large language models (LLMs) have theory of mind? A plethora of papers~and benchmarks have been introduced to evaluate if current models have been able to develop this key ability of social intelligence. However, all rely on limited datasets with simple patterns that can potentially lead to problematic blind spots in evaluation and an overestimation of model capabilities. 
We introduce \methodname{}, the first framework to allow large-scale generation of diverse and challenging theory of mind data for robust training and evaluation.
Our approach leverages an A* search over a custom domain-specific language to produce complex story structures and novel, diverse, yet plausible scenarios to stress test the limits of LLMs. 
Our evaluation reveals that state-of-the-art LLMs, such as Llama-3.1-70B and GPT-4o, show accuracies as low as \finalnumber{0\%} and \finalnumber{9\%} on \methodname{}-generated data, highlighting the need for more robust theory of mind evaluation. As our generations are a conceptual superset of prior work, fine-tuning on our data yields a \finalnumber{27}-point accuracy improvement on the classic ToMi benchmark \citep{le2019revisiting}. 
\methodname{} also enables uncovering underlying skills and factors missing for models to show theory of mind, such as unreliable state tracking or data imbalances, which may contribute to models' poor~performance~on~benchmarks.}

\date{December 12, 2024}
\correspondence{Melanie Sclar at \email{melaniesclar@meta.com}}

\metadata[Code]{\url{https://github.com/facebookresearch/exploretom}}
\metadata[Data]{\url{https://huggingface.co/datasets/facebook/exploretom}}

\begin{document}

\maketitle
\vspace{-0.5em}
\section{Introduction}

Reasoning about other people's intentions, goals, thoughts, and beliefs is a foundation of social intelligence. Known as \textit{Theory of Mind} (ToM)~\citep{premack1978does}, this capability is crucial for effective human interaction. There has been a plethora of recent research that develops theory of mind benchmarks and test LLM capabilities, usually inspired in standard tests for research in children such as the Sally-Anne test~\citep{wimmer1983beliefs}. 
However, these tests are not well-suited for extensively evaluating models, as they focus on specific scenarios and lack the variability and complexity required to remain challenging after online pre-training
. As a result, many existing computational benchmarks may not be effective in robustly evaluating models' theory of mind~abilities. 

\begin{figure}[t]
    \centering
\includegraphics[width=\linewidth]{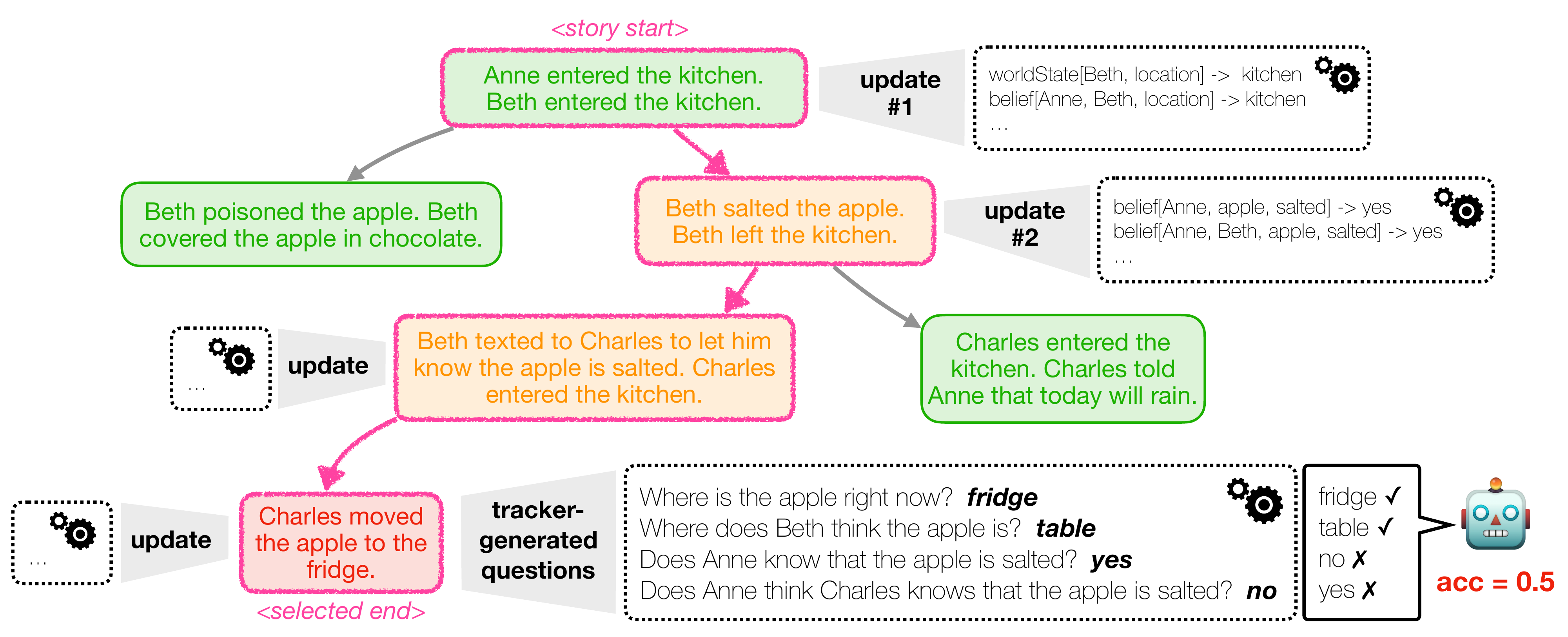}
    \vspace{-2em}
    \caption{\methodname{} finds challenging stories for a given language model by searching through the space of stories supported by its domain-specific language for mental state tracking~(\includegraphics[height=0.8em]{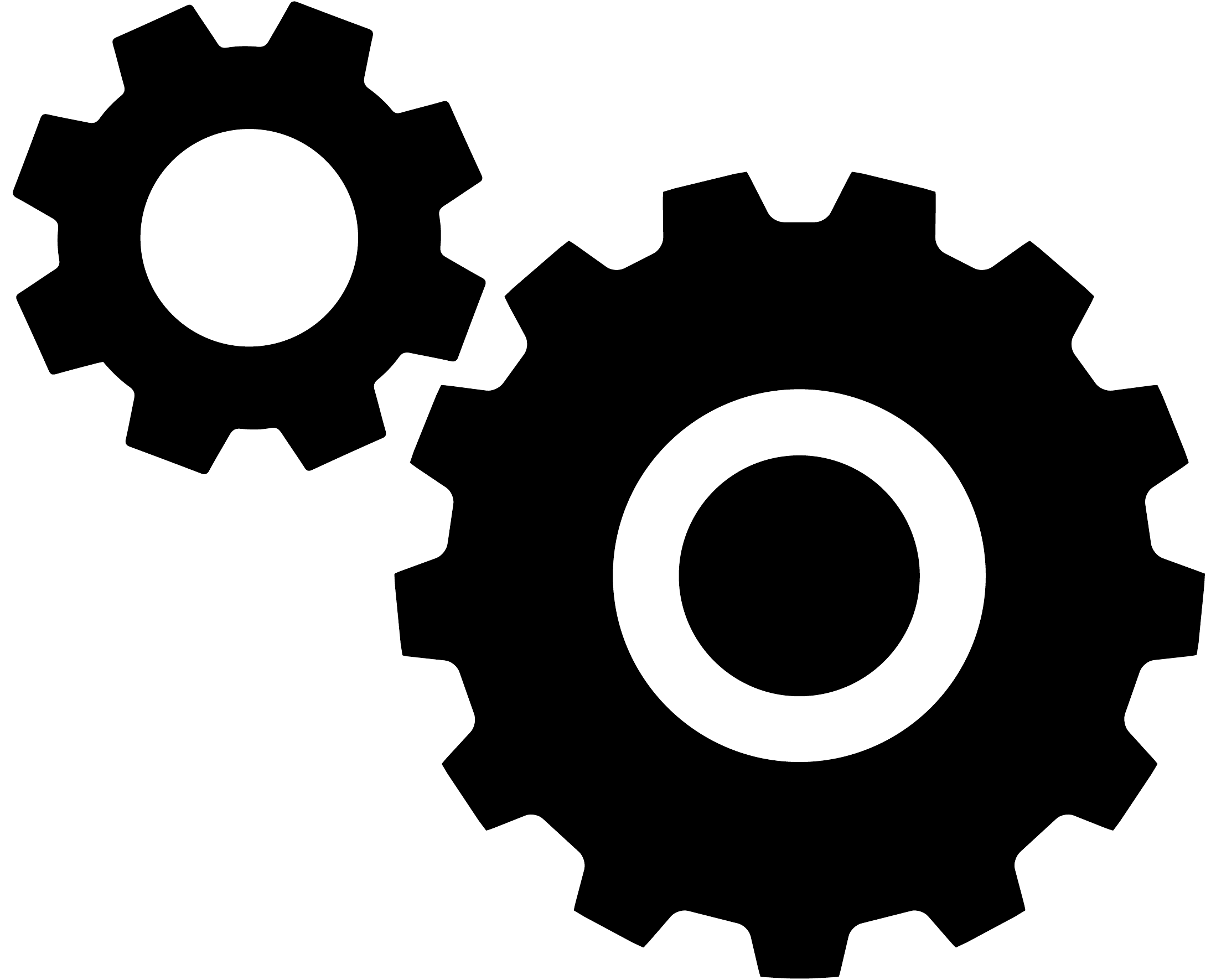}), sampling $k$ supported actions at a time (shown as a node, $k=2$ in the example). Difficulty evaluation (simplified in the figure as \textcolor{easy}{easy}, \textcolor{medium}{medium}, \textcolor{hard}{hard}) of each partial story is done through automatically generated questions
    with reliable ground-truth answers 
    thanks to our tracking procedure. 
    }
    \vspace{-1em}
    \label{fig:dsl}
\end{figure}

We introduce \textbf{\methodname{}}, an A*-powered algorithm for generating reliable, diverse, and challenging theory of mind data that can be effectively employed for testing or fine-tuning LLMs. Our approach leverages a domain-specific language to generate synthetic story structures and their character's  mental states. 
We then use LLMs to create plausible
stories based on these plots, allowing for precise control over the narrative and tracking each character's mental state with high confidence. We employ A* search~\citep{hart1968formal} to efficiently navigate the vast space of possible narratives and pinpoint those that are most likely to fool 
state-of-the-art LLMs. This in turn allows to create a robust, rich 
dataset that effectively tests the limits of current models (Fig.~\ref{fig:dsl}). By generating story structures separately from lexical realizations, we can distinguish the model's core understanding of the social reasoning from vocabulary cues that might give away stylistic hints.

Our contributions are three-fold: we algorithmically address blind spots in theory of mind evaluation, we provide a recipe to create complex training data that helps imbue models with better theory of mind reasoning skills, and we provide insights into why theory of mind skills are still elusive~for~LLMs.

First, our work helps to address the conflicting results from a large number of prior research on evaluating theory of mind~\citep[e.g.][]{sap2022neural, shapira2023clever, fantom, t4d, bigtom, strachan2024naturepaper}, 
including reports that due to oversimplified datasets, prior theory of mind estimates may be overly optimistic~\citep{ullman2023large}. 
Our algorithmic approach also helps address the issue of developing benchmarks that may be close to saturation at the time of release, given the increasingly harder task of anticipating LLM failures \citep[e.g.,][]{biggenbench}. \methodname{}'s adversarial nature allows for generating stories to stress test any LLM, 
diminishing the risk of data leakage onto training data, and thus being more robust than manually-crafted benchmarks. Our experiments show that \textbf{\methodname{}-generated data is extremely challenging, with GPT-4o and Llama-3.1 70B accuracies as low as \finalnumber{9\%} and \finalnumber{0\%} respectively.} \methodname{} supports significantly more scenarios than previously possible, with the unique addition of knowledge gain asymmetry during an interaction, among other improvements.

Second, our method allows creating complex and diverse theory of mind data that can be leveraged for model fine-tuning.
Given theory of mind's implicit nature, it is challenging to find data that explicitly articulates the required reasoning, and existing benchmarks 
are not suitable to use as training data: they 
are often limited in scale~\citep{opentom}, portray specific scripted scenarios~\citep{hitom, le2019revisiting}, and are prone to leakage risks that would make them fully unsuitable for future use. Fine-tuning with this data has been shown to overfit to specific story structures instead of learning the underlying reasoning required 
~\citep{symbolictom}, leading to works focused on creating inference-time algorithms to improve the model's capabilities through prompting or more sophisticated strategies \citep{symbolictom, t4d, simtom, jung2024perceptions}. 
While inference-time methods have proven useful for improving performance in theory of mind benchmarks, 
the benefits of these methods cannot be readily transferred to downstream applications that may also require customized inference-time algorithms for their specific use cases. Fine-tuning Llama-3.1~8B~Instruct on \methodname{}'s data achieves a substantial \textbf{\finalnumber{+27} accuracy point improvement on the classic ToMi benchmark}~\citep{le2019revisiting}, \textbf{showing good generalization to even more complex \methodname{} stories than those seen during training, while still retaining general reasoning capabilities}.

Finally, \methodname{} enables providing new insights into why basic theory of mind reasoning is still challenging for LLMs. 
We show that LLMs struggle with basic state tracking, a fundamental skill underlying theory of mind reasoning: tracking mental states necessarily requires being able to track states
. Our experiments also reveal that in order to improve on theory of mind during fine-tuning, it is crucial to use data that requires theory of mind as opposed to simply requiring state tracking. However, found data (either in-the-wild, or randomly generated) is unlikely to have this necessary property, which may be a key contributor to lagging model performance. 
Overall, \methodname{} offers a valuable tool for  advancing the theory of mind research, enabling the development of more effective LLMs that can better handle complex social interactions.

\section{Adversarially~constructed~stories~with~\methodname{}}\label{sec:storygen}

\begin{figure}[t]
    \centering
\includegraphics[width=\linewidth]{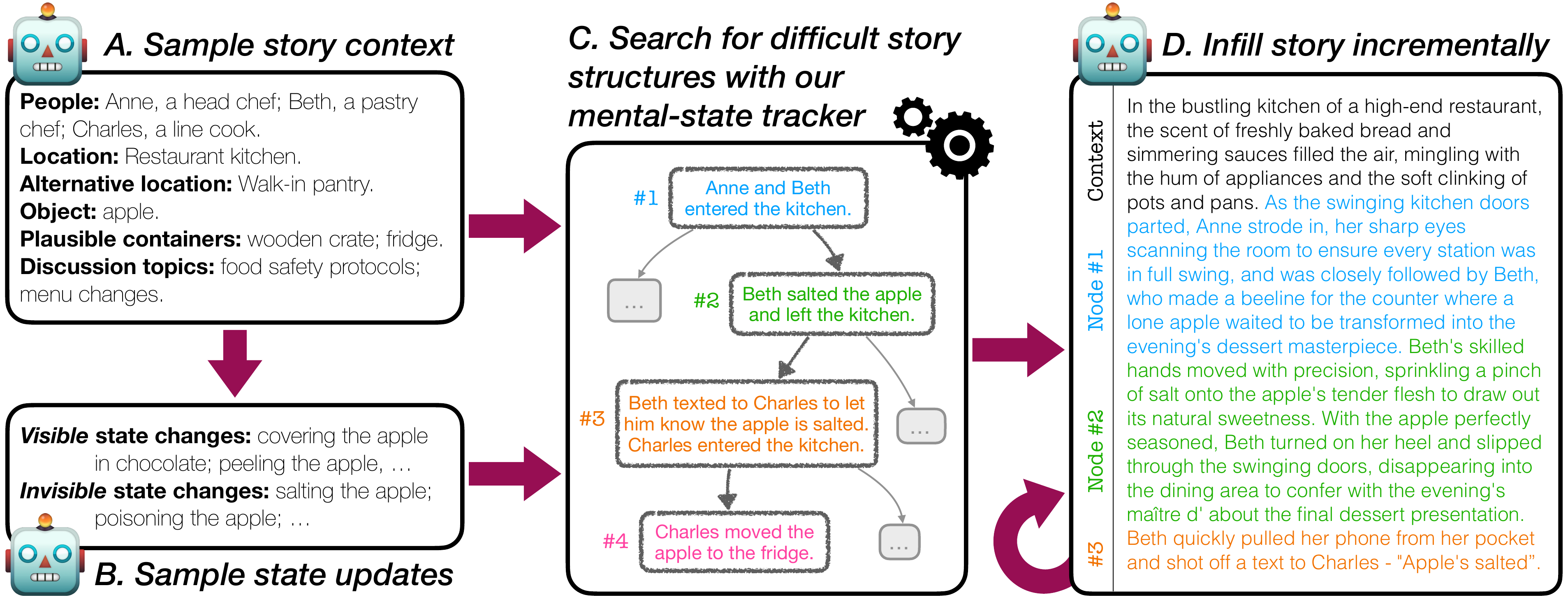}
    \vspace{-1em}
    \caption{Overview of \methodname{}'s story generation procedure. We first sample a plausible story context using an LLM (shown in A and B). Topics discussed, location changes of objects and people, and object state updates, may all be required to track in order to pass our theory of mind tests. We then search for difficult story structures (i.e., the raw story points) by sampling and analyzing different orders in which these actions may be performed using A* search (shown in C, and Fig.~\ref{fig:dsl}). This ensures that the resulting stories will all be challenging tests for models, and may be used for further improvement. Finally, these story structures (nodes \#1-4) are iteratively infilled, one story action at a time, using a language model, yielding a natural-sounding story. Infilled stories are used as training data; benchmarking is done with story structures since they~have~the~highest~reliability.}
    
    \label{fig:flow_example}
\end{figure}


Building on the standard approach in theory of mind of assessing mental state understanding through question answering~\citep{wimmer1983beliefs, kinderman1998theory, baron1999new}, \methodname{} creates stories where different characters may have different beliefs about the current world state and about other people's beliefs, paired with questions to probe model understanding (see Fig.~\ref{fig:dsl}'s highlighted story, along with associated questions probing understanding that e.g. ``Anne does not know that Charles knows that the apple has been salted").

\methodname{}'s story generation process is divided into three main steps: plausible story context sampling (Section~\ref{sec:storysampling}), adversarial story structure generation (Section~\ref{sec:methods_search}), and optionally story infilling (Section \ref{sec:methods_infilling}) -- an example is outlined in Figure~\ref{fig:flow_example}.
We automatically generate questions to probe understanding of said stories as part of the adversarial story structure generation process (Section \ref{sec:methods_question}); this process finds challenging story structures, i.e., story structures that would yield low accuracy with our generated questions. Because questions are generated automatically and directly from the tracked mental and world states, ground truth answers have a high degree of reliability: we do not use language models at all in the question-answer generation procedure.

\subsection{Plausible story context sampling \ \ }
\label{sec:storysampling}
We use an LLM zero-shot to generate a consistent and plausible story context, comprising essential elements such as character names, roles, locations, relevant objects, object containers, and discussion topics (see Fig.~\ref{fig:flow_example}A for a full example). 
This single-step process ensures a coherent and believable setup for our theory of mind stories. 
Previous approaches (such as ToMi \citep{le2019revisiting}) sample objects (e.g., an apple) and object containers independently (e.g. a bottle), often resulting in commonsense violations. Unlike these approaches, our method generates a coherent context by sampling these elements jointly in a single LLM call: autorregresive LLMs will naturally suggest contextually plausible elements based on the ones they already generated, and especially so when explicitly requesting it in the prompt. Additionally, we sample possible object state updates (Figure~\ref{fig:flow_example}B), which are then refined through using an LLM as a judge to filter out implausible and low quality generations. The role of these state updates will be discussed in further detail in Section~\ref{sec:methods_dsl}. The exact prompts used for sampling story contexts are shown in App.~\ref{app:prompts_used}.




\subsection{Adversarially Generating Challenging yet Plausible Story Scripts}\label{sec:methods_search}
\subsubsection{Theory of Mind-Specific Language Definition}\label{sec:methods_dsl}



\methodname{}'s theory of mind-specific language consists of a diverse set of actions $\mathcal{A}$, each transforming the world state and the beliefs of the people involved (the story \textit{state} $s\in\mathcal{S}$). 
A \textit{story} is thus defined as a sequence of actions $(a_1,\ldots, a_n)$, where each action $a_i \in \mathcal{A}$ is a function $a_i: \mathcal{S} \rightarrow \mathcal{S}$. 
Each action also has preconditions to be able to apply it, i.e., restrictions to its domain. For example, a precondition for ``Charles entering the kitchen" is to not be in it already. Applying an action also automatically updates our world state tracking and belief tracking: for example, ``Charles is now in the kitchen"; ``Anne knows that Charles is in the kitchen since they were also in the kitchen"; ``Charles knows that Anne is in the kitchen since he can see her"; and so forth. All these updates and conditions are specifically programmed and tested; see App.~\ref{app:formal}~for~the~full~programs.

\methodname{} enables the generation of diverse stories by significantly 
expanding the range 
of supported actions. These actions include physical changes to the world state such as entering and leaving a room (denoted $a_{\text{enter}}$, $a_{\text{leave}}$), moving an object to a container (or in general, updating its state; denoted $a_{\text{moveObjContainer}}$, $a_{\text{updateObjState}}$ respectively), relocating an object to a different room ($a_{\text{moveObjRoom}}$). Additionally, \methodname{} supports various forms of communication, including: private conversations between two characters, or public broadcasts to all characters in a room; casual discussions about a topic (denoted \textit{chit-chat}), or notifications about changes in the world state (denoted \textit{info}); these actions are referred to as $a_{\text{info-private}}$, $a_{\text{info-public}}$, $a_{\text{chitChat-private}}$, and $a_{\text{chitChat-public}}$. These actions can occur at any point in the story, allowing for a rich and dynamic narrative (see formal definition in App.~\ref{app:formal}) and expanding prior work~\citep{hitom}.

Each new action requires carefully writing the implied belief and world state updates, which precludes scaling the number of actions supported. However, we alleviate this by noting that from a theory of mind perspective, many actions are equivalent. For example, ``peeling an apple" or ``covering an apple in chocolate" have the same implications with respect to belief updates (a \textit{visible property} of the apple is being updated, and the witnesses would be the same). Similarly, poisoning an apple has the same implications as moving an apple from a drawer to a fridge (an \textit{invisible property} is updated, witnesses would be the same, and non-witnesses would not assume there has been an update). The instantiations of these equivalent state updates from a belief perspective are done with an LLM during the story context sampling (see Figure \ref{fig:flow_example}.B).

\vspace{0.6em}
\textbf{Asymmetric belief updates \ \ \ } 
In prior work, all belief updates were \textit{symmetric}: if A and B witnessed an action, then A knows that B witnessed the action and vice versa. Our framework introduces the ability to model asymmetric scenarios. Specifically, we enable the addition of secret witnesses to an action such as someone observing through a security camera, or removal of witnesses without others' knowledge, as in the case of someone becoming distracted by their phone. This added nuance allows for more realistic and complex social scenarios.
Asymmetries $a_{\text{peek}}$ and $a_{\text{distracted}}$ are modifier functions, 
e.g., as a modifier to ``Beth salted the apple"  ($a_{\text{updateObjState}}(\cdot)$) there may be a secret person peeking ($a_{\text{peek}}(a_{\text{updateObjState}}(\cdot))$): 
 ``While this was happening, Diane witnessed it in secret." 

\subsubsection{Generating Questions and Assessing Resulting Story Difficulty}\label{sec:methods_question}
\vspace{-0.3em}
We assess a model's understanding of a generated story $s$=$(a_1, \ldots, a_n)$ 
by probing it with automatically generated question-answer pairs. \methodname{}-generated answers are more reliable than purely-LLM generated ones, since they are directly produced from the states' trajectory with our tracker. 
Questions may be testing first-order beliefs, second-order beliefs, or regular state tracking: \textit{First-order} refers to asking about someone's mental state (e.g., ``Does Anne know the apple is salted?''); \textit{Second-order} refers to one extra level of recursion in mental state tracking (e.g., ``Does Anne think that Charles know the apple is salted?''); \textit{State tracking} may probe about the current state (\textit{ground truth}) or prior ones (\textit{memory}).

We expand the complexity of memory questions with respect to prior work by asking about any intermediate state (e.g. ``Where was the object before X happened?'') instead of solely about the initial one (``Where was the object at the beginning?''). 
Our generated questions are simple to evaluate: they are either binary (yes/no), or are answered by stating an object, container, or room. Specific question formulations differ based on the property, e.g., location (``Where does Charles think that Anne will search for the apple?'') or knowledge (``Does Charles know that the apple is salted?''). See App.~\ref{app:all_questions} for the full list of supported questions.

A question is considered \textit{interesting} if the answer would change depending on the person being asked about. For example ``Does Anne think that Charles knows that the apple is salted?'' is interesting because the answer would differ if asked about someone else, such as ``Does Beth think that Charles knows the apple is salted?''. 
\methodname{}'s tracker very easily allows for automatically detecting interestingness.

\subsubsection{A* Search}
Given a context $\mathcal{C}$ and a set of actions $\mathcal{A}$, our main goal is to find challenging story structures. To increase \methodname{}'s usage flexibility, we support the option of searching for stories $s$ that fulfill desired user conditions 
$\textbf{isDesired}(s) \in \{0, 1\}$
, such as the number of people involved, or the number of actions belonging to a subset $\mathcal{A}'\subseteq\mathcal{A}$ of \textit{important actions}. 

We search over the space of plausible story structures of up to $m$ actions.
We define this space as a directed graph, where each node is a sequence of valid actions $s\!=\!(a_1,\!\ldots,\!a_{i})$, and there is an edge between $s$ and $s'$ if and only if $s$ is prefix of $s'$, and $s'$ contains $k$ more actions than $s$. $k\geq 1$ is the \textit{grouping factor} for actions, defining the granularity with which we will sample and evaluate nodes. For simplicity, Figure~\ref{fig:dsl} depicts only the new $k=2$ actions that each node introduces.

To find challenging stories that simultaneously fulfill the user constraints we use A$^*$~search \citep{hart1968formal}. By definition, A$^*$ selects the path that minimizes $f(s)~=~g(s)~+~h(s)$, where $g(s)$ is the cost of the path from the start to node $s$, and $h(s)$ is a heuristic that estimates the cost of the cheapest path from $s$ to a \textit{goal node} (one of the nodes where it would be acceptable to finish the search). In our context, goal nodes are those such that $\textbf{isDesired}(s') = 1$.
We choose A* as our search algorithm precisely because it enables to search this space prioritizing desired user conditions through $h(s)$, as we will detail below.

A story is said to be challenging for a model if it incorrectly answers our generated questions, i.e., it shows low accuracy.
Thus, we define $g(s)$ as our target model's accuracy among \textit{all} questions for $s$. 
We define the heuristic function $h(s)$ as a proxy estimation of the likelihood of generating a full story $s+s'$ that fulfills user constraints $\textbf{isDesired}(s)=1$, where $s'$ is the continuation of story $s$:
%
%
\begin{align*}
    h(s) = \alpha \Big( 1 - \frac{1}{P}\displaystyle\sum_{i=1}^P \mathbb{1} (\textbf{isDesired}(s + s'_i) = 1)  \Big)
\end{align*}
Here, all $s'_i$ are randomly sampled continuations of $s$ and $0 \leq \alpha \leq 1$ is a scaling factor.
%
A* requires to evaluate all neighbors of a node $s$. Since this would be infeasible given the vast space to explore, and that each $f(\cdot)$ evaluation requires several LLM calls (one per question), we restrict the evaluation to a pre-defined constant number of neighbors, prioritized by the closeness of this node to fulfilling the conditions described by $\textbf{isDesired}(\cdot)$. This pre-defined constant may depend on $f(s)$ to prioritize more promising partial stories (i.e., with lower $f(s)$ values).

\subsection{Story infilling}\label{sec:methods_infilling}

\textit{Story infilling} is the process of transforming a full story structure $s = (a_1, a_2, \ldots, a_n)$ with a story context $\mathcal{C}$ into a natural-sounding narration (see Fig.~\ref{fig:flow_example}D). We infill stories iteratively with an LLM by transforming each action $a$ into a more natural sounding one, according to some stylistic desiderata $d$, and conditioned on the previously infilled context $z$ (denoted $\text{infill}(a, z, d)$). Supported stylistic desiderata $d$ are length requests (e.g., ``use up to two sentences'') or style requests (e.g., ``make this into a conversation''); we optionally also include sampled character goals $g$ and an initial narration context $c$ based on the story $s$, also generated with an LLM (e.g., Anne's goal may be to oversee that all dishes are rapidly delivered to customers; see initial context example in Fig.~\ref{fig:flow_example}). Concretely, the full story infilling $SI$ is as follows:
\begin{align*}
SI(i)=\text{infill}(a_i, SI(i-1), d_i, g) \text{ where } SI(0)=c
\end{align*}
%
Infilling is done iteratively to ensure that the order of the actions stays the same, since this is important for keeping the mental state tracking valid. To further increase reliability, we use an LLM as a judge after each infilling step to confirm that each mental state tracked after executing the story step $a_i$ still holds even after infilling. This discards infillings that introduced ambiguity or hallucinations.

\section{\methodname{} as an evaluation benchmark}\label{sec:ttm_eval}
\label{eval_benchmark}
We begin by showcasing how \methodname{} story structures can be used as a challenging benchmark, highlighting its unique features and advantages.  

\textbf{Experimental setup \ \ \ }
We use \methodname{} to generate \finalnumber{10} story structures for each of \finalnumber{9} action sets (each with and without asymmetry) and each set of user conditions. Each story generation is allowed to evaluate \finalnumber{50} nodes. User conditions---$\text{isDesired}(\cdot)$---require exactly $p\in\{2,3,4\}$ people involved, with $a\in\{2,3,4\}$ actions belonging to the set of important actions $\mathcal{A}'$
, spanning across either $r=1$ or $r=2$ rooms, and with $m\leq15$ actions in total---leading to a total of 162 settings. In all experiments, $\mathcal{A}'$ are the actions that add new basic world knowledge: 
$\mathcal{A}'\!=\!\{a_{\text{moveObjContainer}}, a_{\text{updateObjState}}, a_{\text{moveObjRoom}}, a_{\text{chitChat-*}}\}$. We then infill every story. We use Llama-3.1-70B-Instruct~\citep{llama3}, GPT-4o~(\cite{gpt4o}; queried early Dec. 2024), and Mixtral-8x7B-Instruct~\citep{mixtral} to generate story structures. A* is run with $\alpha=0.1$, $P=50$, and $k=3$ (i.e. grouping three actions per node). See generation examples in App.~\ref{app:generation_examples}.

\begin{table}[t]
\centering
\caption{Accuracy results of  \methodname{}'s story structures on 18 action sets $\mathcal{A}$, each aggregating 90 total stories from 9 different settings (number of people, actions, and rooms). Each set is either based on actions supported by well-known theory of mind tests or includes our novel expansions, and is analyzed excluding or including asymmetry (\xmark{}, \cmark{}). Each setting requires at least one action in the story to be from one of the \boxed{\text{squared}} actions to encourage non-overlapping story structure characteristics between action sets shown. Data was generated using each model as its own evaluator (i.e., as $g(\cdot)$), and results shown include all first-order questions---the most basic theory of mind level, not requiring recursion. Lowest accuracy for each model is bolded.
}\label{tab:eval_benchmark_accuracy}
\begin{tabular}{@{}ccccccc@{}}
\toprule
\begin{tabular}[c]{@{}c@{}}\methodname{} action set \\ $\{a_{\text{enter}}, a_{\text{leave}}, \ldots$\end{tabular} & \multicolumn{2}{c}{\begin{tabular}[c]{@{}c@{}}Llama-3.1 \\ 70B Inst.\end{tabular}} &
\multicolumn{2}{c}{\begin{tabular}[c]{@{}c@{}}GPT-4o\end{tabular}} &
\multicolumn{2}{c}{\begin{tabular}[c]{@{}c@{}}Mixtral\\8x7B Inst.\end{tabular}} \\ \midrule
include asymmetry modifiers? ($a_{\text{peek}}$, $a_{\text{distracted}}$) & \xmark & \cmark
 & \xmark & \cmark
 & \xmark & \cmark
 \\ \midrule \toprule

$\ldots, \boxed{a_{\text{moveObjContainer}}}\}$ &  .18 & \textbf{.00} &  .40 & .25 & .37 & .33 \\
$\ldots, \boxed{a_{\text{updateObjState}}}\}$  & .27 & .24 & .25 & .24 & .03 & .01  \\
$\ldots, \boxed{a_{\text{moveObjContainer}}}, \boxed{a_{\text{updateObjState}}}\}$  &  .26 & .03 & .35 & .31  &  .24 & .09 \\
$\ldots,a_{\text{moveObjContainer}}, \boxed{a_{\text{moveObjRoom}}}\}$  &  .11 & .10 & \textbf{.09} & \textbf{.16} & \textbf{.00} & \textbf{.00} \\
$\ldots, a_{\text{moveObjContainer}}, \boxed{a_{\text{info-*}}}\}$  & \textbf{.06} & .07 & .29 & .25 &  .35 & .36  \\
$\ldots, a_{\text{moveObjContainer}}, a_{\text{moveObjRoom}}, \boxed{a_{\text{info-*}}} \}$  & .11 & .07 & .24 & .24 & .03 & .04 \\
$\ldots, a_{\text{moveObjContainer}}, a_{\text{moveObjRoom}}, a_{\text{chitChat-*}}, \boxed{a_{\text{info-*}}} \}$  & .72 & .69 & .73 & .68  &  .53 & .47 \\
$\ldots, \boxed{a_{\text{chitChat-private}}} \}$  & .75 & .66 & .77 & .59 & .51 & .45   \\
$\ldots, \boxed{a_{\text{chitChat-public}}} \}$  & .60 & .55 & .49 & .47 & .37 & .34  \\
\bottomrule
\end{tabular}
\end{table}

\begin{figure}[t]
\includegraphics[width=0.49\textwidth]{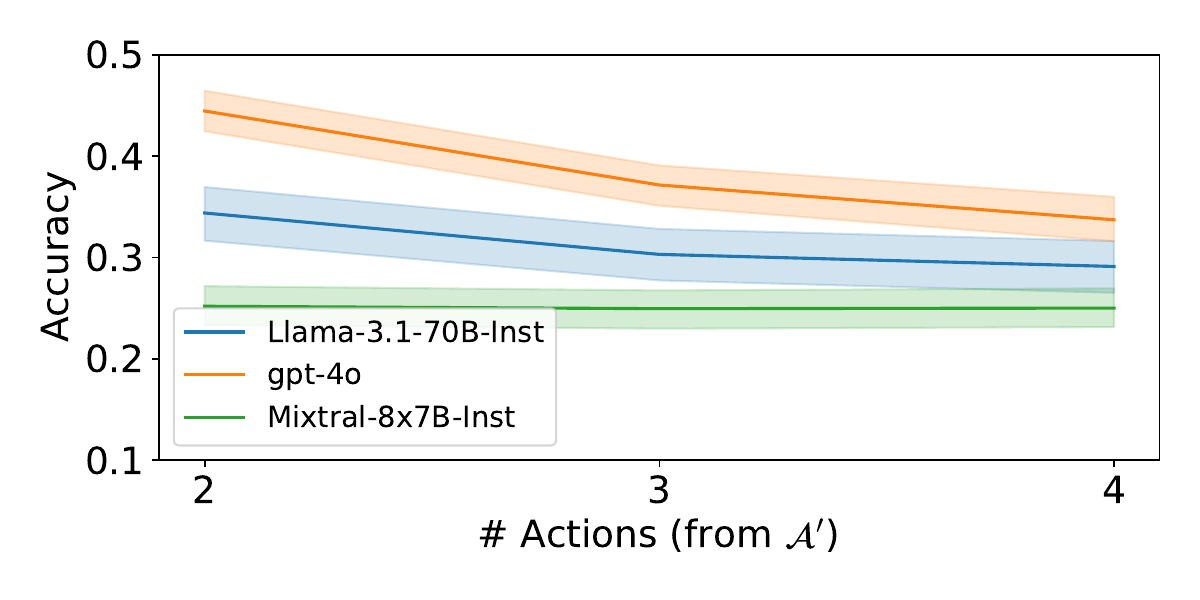} 
\includegraphics[width=0.49\textwidth]{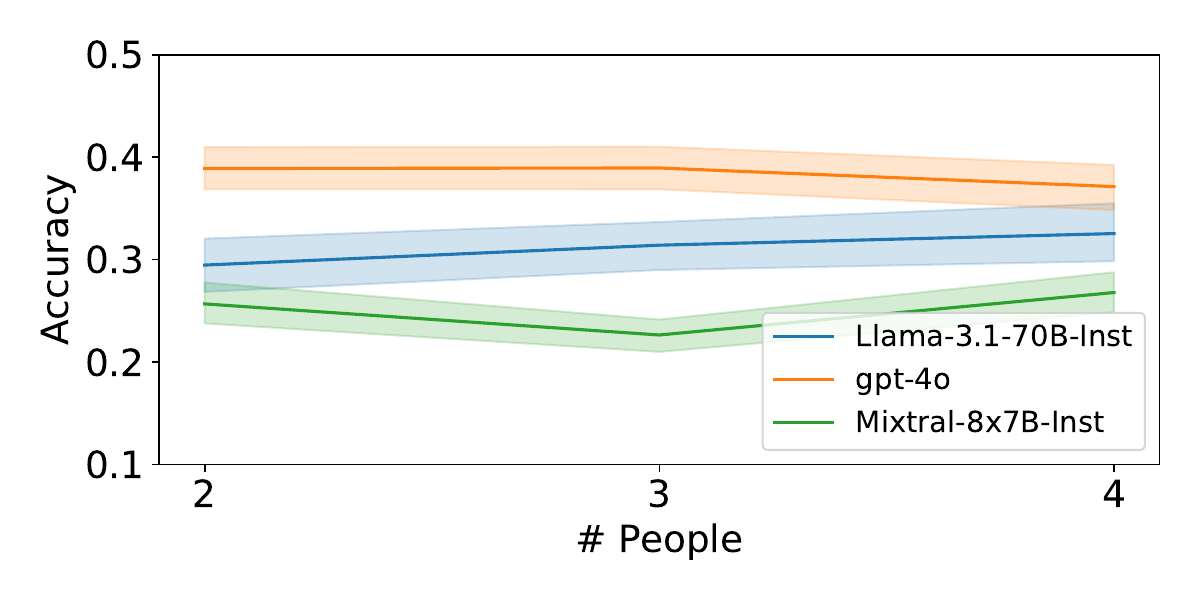}
\caption{Accuracy on \methodname{}'s story structures when increasing the number of actions or people involved. Accuracy is computed across all story structure settings. Difficulty of \methodname{}-generated stories tends to increase or stay similar when increasing the number of actions. A story with greater number of people suggests similar or lower difficulty, possibly because when fixing the number of actions there are fewer actions per person (see details in App.~\ref{app:counterintuitive_phenomenon}). 
}\label{fig:num_moves_people}
\end{figure}

\textbf{\methodname{} finds challenging story structures for frontier models \ \ \ } 
As shown in Table~\ref{tab:eval_benchmark_accuracy}, our \methodname{} consistently identifies story structures that are highly challenging for models across various action sets, with average performances in \methodname{}-generated datasets as low as \finalnumber{0.09} for GPT-4o (i.e., \finalnumber{9\%}). {When increasing the number of actions, difficulty tends to increase or remain similarly challenging. Performance tends to stay the same or decrease when increasing the number of people involved, possibly because with a fixed number of state-changing actions there will be fewer actions per person which may be easier to track. 
See Figure~\ref{fig:num_moves_people} and App~\ref{app:counterintuitive_phenomenon}.

%
%

\textbf{A* is a better strategy than over-generation and filtering \ \ \ } Over-generation and filtering has become a standard procedure for synthetic data generation \citep[e.g.][]{west-etal-2022-symbolic, wang-etal-2023-self-instruct}
. We measure the effectiveness of A* by comparing the A*-generated data to the data resulting from over-generating stories with our domain-specific language---using the same $\textbf{isDesired}(\cdot)$ criteria and budget as used in the A* search---and retaining only the most difficult stories.
In a set of \finalnumber{81} randomly-selected settings (50\% of the original 162 settings, due to the experiment's high cost), we generate 50 stories with each method using Llama-3.1-70B-Instruct and a budget of 2500 accuracy evaluations each.
A* yielded a more challenging dataset (by 2 accuracy points), with shorter stories on average (1.6 fewer actions). This length difference is possibly due to the pressures A* induces towards shorter stories through the heuristic $h(s)$. 
See Figure~\ref{app:astar_vs_baseline} for the full distribution of results.

\textbf{Story structures found adversarially for a model remain challenging for other models \ \ } We evaluate the difficulty of a \methodname{}-generated dataset with each model, and find that although there is an increased difficulty towards data generated adversarially with the same model, it remains challenging for all others. Notably, the generated datasets remain challenging even when adding question types not included in the $g(\cdot)$ optimization (second-order belief questions). See Table~\ref{tab:crossmodeleval}.

\begin{table}[t]
\centering
\caption{Accuracy 
results on \methodname{}-generated data built to minimize accuracy for each particular model. A random sample of 1000 (story structure, question) pairs is shown. Data remains challenging even if it was built with a different model, and even including questions we did not optimize for: story structures were selected adversarially towards first-order belief questions only ($g(\cdot)$), accuracies shown include all belief questions.}\label{tab:crossmodeleval}
\begin{tabular}{@{}cccc@{}}
\toprule
 & \multicolumn{3}{c}{Model used for evaluation}        \\ \midrule
\begin{tabular}[c]{@{}c@{}}Model used in \methodname{}  \\ generation ($g(\cdot)$) \end{tabular} & \begin{tabular}[c]{@{}c@{}}Llama-3.1\\ 70B Inst.\end{tabular} & GPT-4o & \begin{tabular}[c]{@{}c@{}}Mixtral \\ 7x8B Inst.\end{tabular} \\ \midrule \toprule
Llama-3.1 70B Inst.  &  0.57 & 0.68 & 0.32 \\ 
GPT-4o    & 0.60 & 0.61 & 0.32  \\ 
Mixtral 7x8B Inst.  & 0.68 & 0.74 & 0.30 \\ \bottomrule
\end{tabular}
\end{table}

\textbf{Humans agree with \methodname{}-generated story structures labels \ \ \ }
We conducted a human evaluation to verify the quality of the story structures' automatically-generated labels and the story infillings.
For labels, we annotated \finalnumber{100} questions across \finalnumber{12} randomly-sampled story structures from all settings generated for Table~\ref{tab:eval_benchmark_accuracy}, 
and found \finalnumber{99\%} agreement with our expected answers---likely due to the clear and concise nature of our stories and that 
the ground truth labels were generated by our domain-specific language. We measure story infilling quality by repeating the question-answering procedure with a different set of \finalnumber{100} questions across \finalnumber{12} randomly-sampled infilled story structures. In this case, the human agreed with the ground truth label \finalnumber{89\%} of the time---a small degradation likely due to the LLM-powered method introducing ambiguity. 


\textbf{Infilled stories remain challenging \ } Infilled stories with Llama-3.1 70B yielded an average accuracy of \finalnumber{0.61}. Although the average accuracy increased by \finalnumber{0.12} through the infilling process, the samples remained challenging thanks to the highly challenging underlying stories\footnote{Infilling can be also added to the A* search; we deemed it unnecessary given that this simpler method still yields a highly challenging benchmark and it is less costly.}. One key factor for this accuracy difference comes from models sometimes making the mental states more explicit through the infilling process: results shown correspond to a single attempt at infilling each story (41\% of the samples ended successfully in a single attempt, judged by an LLM). Although stories remain challenging, since infilling with an LLM may introduce some ambiguities or hallucinations we only use them as training data. 
See App.~\ref{app:infilling} for detailed results for all action sets.

\section{\methodname{} is effective as training data generator}\label{sec:ttm_train}

\textbf{Experimental setup \ \ }
We fine-tune Llama-3.1 8B Instruct using a dataset of \finalnumber{79700} (story, question, answer) triples, focusing solely on the completion tasks, and dub the resulting model \methodname{}-8B. The dataset comprises both raw story structures and infilled stories, incorporating story structures from each of the \finalnumber{9} action sets listed in Table~\ref{tab:eval_benchmark_accuracy} (excluding asymmetry, and with a balanced number of questions per story type), and various user constraints---the same as in Section~\ref{sec:ttm_eval}. 
We do full fine-tuning with the following hyperparameters: a learning rate of $10^{-6}$, 100 warm-up steps, effective batch size of 40 samples, where we fine-tune solely on completions.

\textbf{Fine-tuning with \methodname{} generalizes well to \methodname{}-generated data with more people and more actions than used in training \ \ \ } Since \methodname{}-8B is trained with \methodname{}-generated data involving $p=\{2, 3, 4\}$ people with  $m=\{2, 3, 4\}$ actions from the set of important actions $\mathcal{A}'$, we evaluate generalization within the \methodname{} domain by evaluating on \methodname{}-generated data involving 5 people, and up to 11 actions. This data is generated with Llama-3.1, the same model as original training data. See Figure~\ref{fig:generalization-in-ttm}.

\begin{table}[t]
\caption{Performance on major false-belief benchmarks; accuracy (in \%) unless otherwise stated: OpenToM uses F1 score. Parenthesis reflect differences
between out-of-the-box model and fine-tuned version using \methodname{}-generated data. \textbf{Bold} reflects higher overall performance.}\label{tab:tom_benchmarks_acc}
\centering
\begin{tabular}{@{}cccccc@{}}
\toprule
                  & ToMi & Hi-ToM & BigToM & OpenToM (F1) & FANToM \\ \midrule
Llama-3.1 8B Instruct & \finalnumber{68\%}  &   \finalnumber{30\%}     &    \finalnumber{75\%}    &    \finalnumber{.39}     &    \finalnumber{0.3\%}  \\   
\methodname{}-8B   &   \finalnumber{\textbf{95\%} (+27\%)}   &    \finalnumber{\textbf{59\%} (+29\%)}    &    \finalnumber{\textbf{81\%} (+6\%)}    &   \finalnumber{\textbf{.46} (+.07)}      &    \finalnumber{0.2\% (-0.1\%)}    \\  \bottomrule  
\end{tabular}
\end{table}

\begin{figure}[t]
\centering
    \begin{minipage}{0.47\textwidth}
    \includegraphics[width=\linewidth]{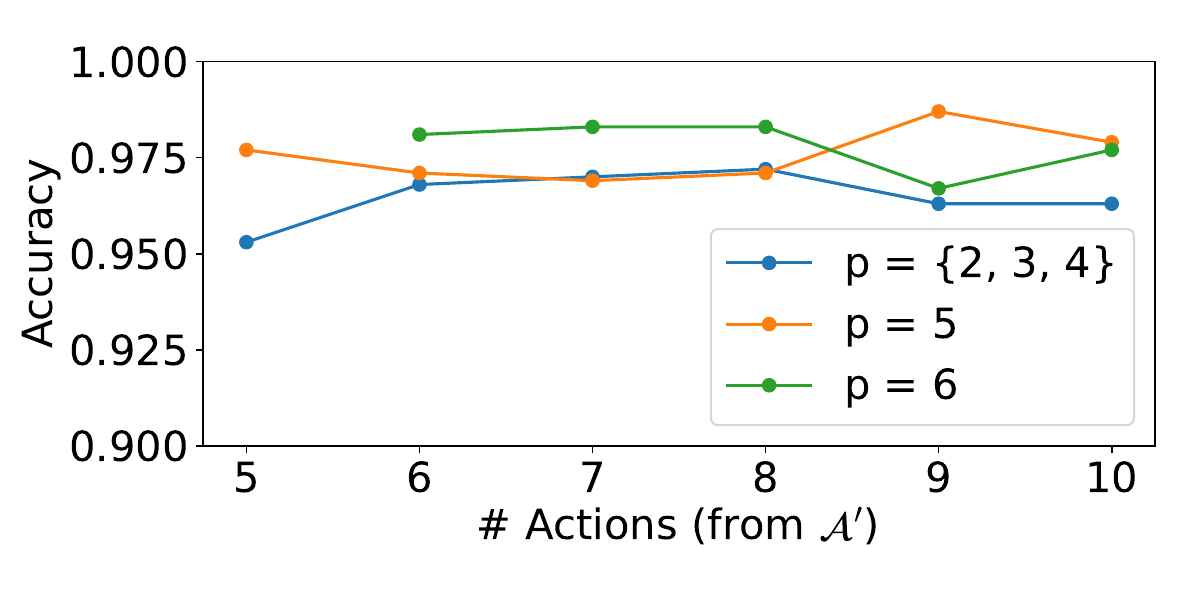}
    \vspace{-2em}
    \caption{\methodname{}-8B accuracy when evaluating on \methodname{}-generated data with more people $p$ and/or more actions $a$ than seen during training ($p\!<\!5, a\!<\!5$). Performance remains high when adding several actions and/or up to two people.
    }
    \label{fig:generalization-in-ttm} 
    \end{minipage}\hfill
\begin{minipage}{0.47\textwidth}
  \centering
    \includegraphics[width=\textwidth]{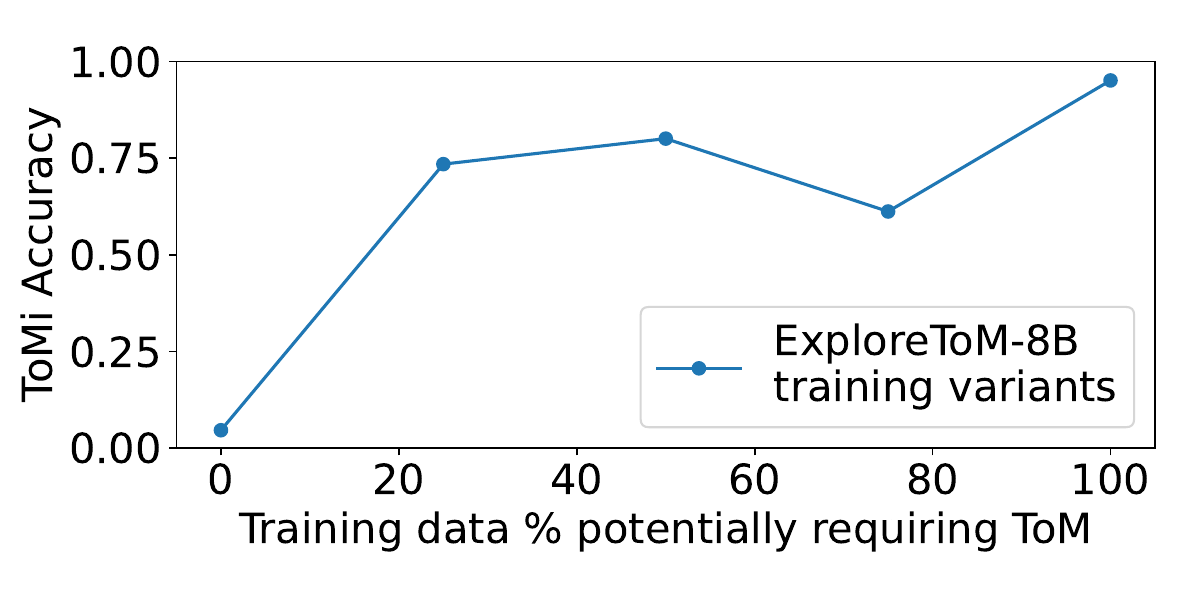} 
    \vspace{-2em}
    \caption{ToMi accuracy when training with \methodname{}-generated data with different proportions of interesting questions (i.e., questions potentially requiring theory of mind to answer). Here, all variants are fine-tuned with \finalnumber{85000} story structure samples for 1 epoch. 
    }\label{fig:training_mixture}  
\end{minipage}
\end{figure}

\textbf{Fine-tuning with \methodname{} improves or maintains performance on theory of mind benchmarks without hurting general reasoning capabilities \ \ }
We evaluate our fine-tuned \methodname{}-8B model on five prominent theory of mind benchmarks: ToMi~\citep{le2019revisiting}, Hi-ToM~\citep{hitom}, BigToM~\citep{bigtom}, OpenToM~\citep{opentom}, and FANToM~\citep{fantom}. Results show significant improvements in performance on ToMi and HiToM, with accuracy gains of \finalnumber{+27} points on both benchmarks (see Table~\ref{tab:tom_benchmarks_acc}). The model maintains or shows small gains on the remaining three similar benchmarks, indicating that fine-tuning on \methodname{} data enhances or preserves performance across a range of theory of mind tasks\footnote{The lack of performance difference in FANToM is likely due to its length confounding factor: data points can have 1000+ tokens, yet our model is fine-tuned with a maximum length of 300 tokens.}.



We also evaluate out-of-domain reasoning skills using the two datasets: Multi3Woz \citep{multi3woz}, a commonly-used dataset for dialogue state tracking, and MMLU~\citep{mmlu}, which tests both world knowledge and problem-solving abilities. Dialogue state tracking capabilities are preserved: both the base model and \methodname{}-8B achieve \finalnumber{96\%}. Broader reasoning capabilities are also generally preserved, with a small \finalnumber{2\%} performance difference (base model achieves 69\%; \methodname{}-8B, \finalnumber{67\%}). Given the out-of-domain nature, we expect that intermixing data with samples more similar to MMLU's domains 
will substantially alleviate this slight regression.


\textbf{Data mixture affects downstream performance \ \ \ }
We fine-tune five models, with 0\%, 25\%, 50\%, 75\%, or 100\% of the stories requiring theory of mind to answer at least one question about the story. Figure~\ref{fig:training_mixture} shows that training with as much stories that require theory of mind is crucial for achieving high downstream performance (using ToMi as a proxy dataset), even if some of the individual questions used for training do not require theory of mind.

\section{On underlying skills needed for theory of mind}\label{skills_of_tom}

\methodname{} enables uncovering and quantifying underlying causes for models' poor theory of mind reasoning in models out-of-the-box. We specifically focus on the lack of robust state tracking skills, and the need for targeted training data in order to improve theory of mind capabilities.

%
%
%
\textbf{LLMs lack robust state tracking skills \ \ \ } \methodname{}'s objective is to find story structures where models fail to answer questions; some of these questions simply require state tracking, specifically the ones where every person would give the same answer (i.e., their mental state is the same in this regard; e.g., in Fig.~\ref{fig:dsl}, all $X\in\{\text{Anne, Beth, Charles}\}$ would answer the same to ``Where does X think Anne is right now?''). 
By definition (see \S~\ref{sec:methods_question}), these are the \textit{uninteresting} questions. \methodname{}-generated questions are approximately evenly split between interesting and uninteresting, and uninteresting ones are even more challenging on average: the accuracy of interesting and uninteresting questions is 49\% and 31\% respectively for Llama-3.1 70B, 58\% and 37\% for GPT-4o, and 45\% and 26\% for Mixtral. See Table~\ref{tab:int_vs_no_int} in App.~\ref{app:int_vs_no_int} for full breakdown for all settings.

State tracking questions are a subset of theory of mind questions, and arguably an easier case since the required logic for answering questions is simpler. Therefore, improving models' performance on state tracking may be a crucial prerequisite for achieving theory of mind reasoning in LLMs. As we have demonstrated, \methodname{} can be easily adapted to stress test pure state tracking, simply by retaining only the uninteresting questions.

\textbf{Training data biases against theory of mind and its implications \ \ \ } Figure~\ref{fig:training_mixture} shows that to successfully improve performance on the ToMi benchmark, \methodname{} fine-tuning data needs to be biased towards \textit{interesting} questions. However, a significant portion of models' training data is likely biased against requiring the tracking of divergent mental states (e.g., news articles).

As a conceptual proof that this phenomena occurs even within our custom domain-specific language unless we explicitly bias towards theory of mind, we demonstrate that randomly-sampled story structures tend not to require theory of mind. Using \methodname{}'s domain-specific language, we randomly generate
\finalnumber{1000} story structures with ToMi primitives (\{$a_{\text{enter}}$, $a_{\text{leave}}$, $a_{\text{moveObjContainer}}$\}) for stories involving $\{2, 3, 4\}$ people and $\{2, 3, 4\}$ object movements. 
We consider a story to not require theory of mind if all first-order and second-order theory of mind questions are \textit{un-interesting}, as defined in \S~\ref{sec:methods_question} (i.e., all share the same mental state). This stringent criterion evaluates \textit{all} questions simultaneously. Nevertheless, our results show that 78\% or more of the randomly-sampled stories meet this condition across all settings, with up to 87\% of stories fulfilling the condition for the smallest setting (2 people, 2 object movements). When considering each question individually, 91\%-95\% are uninteresting questions. See App.~\ref{app:random_sample_stories} for more details.


\section{Related Work}
\vspace{-0.3em}

\textbf{Theory of mind benchmarking for language models \ \ \ } 
Theory of mind benchmarks in language models can be categorized into human-generated and model-generated datasets. While human-generated datasets ~\citep{fauxpas, biggenbench, tombench} test reasoning about goals, emotions of others, and future actions, they are often limited in size and scope. 
Machine-generated datasets, such as foundational ToMi~\citep{le2019revisiting} and its successor Hi-ToM~\citep{hitom} focus primarily on mental state tracking, but have significant limitations: ToMi only supports a restricted set of actions (\{$a_{\text{enter}}$, $a_{\text{leave}}$, $a_{\text{moveObjContainer}}$\}), while Hi-ToM adds $a_{\text{info-*}}$ but only as the last action in a story, and both datasets have extremely restricted interactions to orders. In contrast, our method, \methodname{}, significantly expands the scope of machine-generated datasets by supporting a larger number of actions, diverse wording, and plausible contexts. Unlike recent approaches that rely on LLMs for generation~\citep{fantom,opentom,bigtom}, \methodname{} ensures reliability and multi-interaction storytelling, making it a more comprehensive and robust benchmark for theory of mind in LLMs.
\textbf{Theory of mind beyond language modeling} 
Theory of mind has been explored in various areas, including human computer interaction~\citep{wang2021hcitom}, explainable AI~\citep{AKULA2022103581}, and multi-agent reinforcement learning \citep{rabinowitz2018machine, sclar22symmetric, zhu2021few}. Recent benchmarks have evaluated theory of mind in multi-modal settings~\citep{mmtomqa} and multi-agent collaboration \citep{bara2021mindcraft, shi2024muma}, but these focus on goal-driven interactions. Psychologists distinguish between affective (emotions, desires) and cognitive (beliefs, knowledge) theory of mind~\citep{shamay2010role}, with cognitive theory of mind developing later in children~\citep{wellman2014making}. Our work targets cognitive theory of mind, which is well-suited for generating situations with a domain-specific language and provides unambiguous answers across cultures. By focusing on cognitive theory of mind, our approach complements existing research and provides a comprehensive benchmark for this crucial aspect of human reasoning in language models. 

\textbf{Synthetic data generation}
Synthetic data has become promising approach for acquiring high-quality data in various domains, including multihop question-answering~\citep{lupidi2024source2synth}, and language model evaluation~\citep{wang2024self}. The process involves data augmentation/generation and curation, with techniques such as permutation-based augmentation~\citep{yu2023metamath, li2024gsm} and  iterative prompting~\citep{yang2022re3}. However, model hallucination \citep{guarnera2020deepfake, van2023synthetic, wood2021fake, zhang2023siren} requires careful filtration and curation to ensure data quality. While prior works have used external feedback~\citep{zelikman2022star, luo2023wizardcoder}, our approach leverages an external LLM-as-judge to evaluate the plausibility and challenge of generated stories, both before and after infilling. Recently, AutoBencher~\citep{autobencher} has also been proposed to automatically search for datasets that meet a salience, novelty, and difficulty desiderata, highlighting the importance of careful benchmark creation. 
%
%
Unlike AutoBencher, which over-generates under the assumption that text-based conditioning minimizes hallucinations, our approach lifts this assumption and actively searches the space of possible narratives. This enables to create high-quality synthetic data regardless of the likelihood of a story being generated zero-shot, and generating even more challenging stories than with over-generation.


%
\vspace{-0.5em}
\section{Conclusions}
\vspace{-0.3em}
Theory of mind (ToM) is essential for social intelligence, and developing agents with theory of mind is a requisite for efficient interaction and collaboration with humans. Thus, it is important to build a path forward for imbuing agents with this type of reasoning, as well as methods for robustly assessing the  of models' theory of mind reasoning capabilities.

We present \methodname{}, an A*-powered algorithm for generating reliable, diverse and challenging theory of mind data; specifically, creating synthetic stories that require theory of mind to understand them, along with questions to probe understanding.
\methodname{}'s adversarial nature enables the stress testing of future models and making our evaluation more robust to data leakage. We show that \methodname{} generates challenging theory of mind evaluation sets for many frontier models, with accuracies as low as \finalnumber{0\%} for Llama-3.1 70B Instruct and \finalnumber{9\%} for GPT-4o. Moreover, we show that \methodname{} can be used as a method for generating training data, leading to improvements of up to \finalnumber{29} accuracy points in well-known theory of mind benchmarks. Synthetic data is crucial for this domain, given that data that articulates theory of mind reasoning is difficult to find in the wild: children have access to a wide range of naturalistic social settings that incentivize the development of theory of mind but there is no such parallel pressure for LLMs.


Finally, we provide insights as to why basic theory of mind is still elusive to LLMs, including poor state tracking skills and demonstrating the need for training data that purposefully requires theory of mind, which is likely not present in the wild nor in randomly-generated data.



\section*{Limitations}
\methodname{} offers a valuable tool for theory of mind research, and is a first step towards developing LLMs that can handle social interactions effectively. Although its data encompasses diverse and challenging settings---more than previously available---, and is grounded in established psychological tests, \methodname{} necessarily simplifies the complexity of real-world states and narratives by constraining it to the supported types of actions and interactions. Our framework requires manual coding of new actions, wich can be time-consuming process but comes with the benefit of a significant reliability improvement. Furthermore, our stories are not necessarily goal-oriented narratives, highlighting an important avenue for future work: creating datasets where actions stem directly from character goals to further enhance diversity and plausibility.

\clearpage
\newpage
\bibliographystyle{assets/plainnat}
\bibliography{paper}

\begin{thebibliography}{51}
\providecommand{\natexlab}[1]{#1}
\providecommand{\url}[1]{\texttt{#1}}
\expandafter\ifx\csname urlstyle\endcsname\relax
  \providecommand{\doi}[1]{doi: #1}\else
  \providecommand{\doi}{doi: \begingroup \urlstyle{rm}\Url}\fi

\bibitem[Akula et~al.(2022)Akula, Wang, Liu, Saba-Sadiya, Lu, Todorovic, Chai, and Zhu]{AKULA2022103581}
Arjun~R. Akula, Keze Wang, Changsong Liu, Sari Saba-Sadiya, Hongjing Lu, Sinisa Todorovic, Joyce Chai, and Song-Chun Zhu.
\newblock Cx-tom: Counterfactual explanations with theory-of-mind for enhancing human trust in image recognition models.
\newblock \emph{iScience}, 25\penalty0 (1):\penalty0 103581, 2022.
\newblock ISSN 2589-0042.
\newblock \doi{https://doi.org/10.1016/j.isci.2021.103581}.
\newblock \url{https://www.sciencedirect.com/science/article/pii/S2589004221015510}.

\bibitem[Bara et~al.(2021)Bara, Sky, and Chai]{bara2021mindcraft}
Cristian-Paul Bara, CH-Wang Sky, and Joyce Chai.
\newblock Mindcraft: Theory of mind modeling for situated dialogue in collaborative tasks.
\newblock In \emph{Proceedings of the 2021 Conference on Empirical Methods in Natural Language Processing}, pages 1112--1125, 2021.

\bibitem[Baron-Cohen et~al.(1999)Baron-Cohen, O’Riordan, Jones, Stone, and Plaisted]{baron1999new}
Simon Baron-Cohen, Michelle O’Riordan, Rosie Jones, Valerie Stone, and Kate Plaisted.
\newblock A new test of social sensitivity: Detection of faux pas in normal children and children with asperger syndrome.
\newblock \emph{Journal of Autism and Developmental Disorders}, 29\penalty0 (5):\penalty0 407--418, 1999.

\bibitem[Chen et~al.(2024)Chen, Wu, Zhou, Wen, Bi, Jiang, Cao, Hu, Lai, Xiong, et~al.]{tombench}
Zhuang Chen, Jincenzi Wu, Jinfeng Zhou, Bosi Wen, Guanqun Bi, Gongyao Jiang, Yaru Cao, Mengting Hu, Yunghwei Lai, Zexuan Xiong, et~al.
\newblock Tombench: Benchmarking theory of mind in large language models.
\newblock \emph{arXiv preprint arXiv:2402.15052}, 2024.

\bibitem[Dubey et~al.(2024)Dubey, Jauhri, Pandey, Kadian, Al-Dahle, Letman, Mathur, Schelten, Yang, Fan, et~al.]{llama3}
Abhimanyu Dubey, Abhinav Jauhri, Abhinav Pandey, Abhishek Kadian, Ahmad Al-Dahle, Aiesha Letman, Akhil Mathur, Alan Schelten, Amy Yang, Angela Fan, et~al.
\newblock The llama 3 herd of models.
\newblock \emph{arXiv preprint arXiv:2407.21783}, 2024.

\bibitem[Gandhi et~al.(2024)Gandhi, Fr{\"a}nken, Gerstenberg, and Goodman]{bigtom}
Kanishk Gandhi, Jan-Philipp Fr{\"a}nken, Tobias Gerstenberg, and Noah Goodman.
\newblock Understanding social reasoning in language models with language models.
\newblock \emph{Advances in Neural Information Processing Systems}, 36, 2024.

\bibitem[Guarnera et~al.(2020)Guarnera, Giudice, and Battiato]{guarnera2020deepfake}
Luca Guarnera, Oliver Giudice, and Sebastiano Battiato.
\newblock Deepfake detection by analyzing convolutional traces.
\newblock In \emph{Proceedings of the IEEE/CVF conference on computer vision and pattern recognition workshops}, pages 666--667, 2020.

\bibitem[Hart et~al.(1968)Hart, Nilsson, and Raphael]{hart1968formal}
Peter~E Hart, Nils~J Nilsson, and Bertram Raphael.
\newblock A formal basis for the heuristic determination of minimum cost paths.
\newblock \emph{IEEE transactions on Systems Science and Cybernetics}, 4\penalty0 (2):\penalty0 100--107, 1968.

\bibitem[Hendrycks et~al.(2021)Hendrycks, Burns, Basart, Zou, Mazeika, Song, and Steinhardt]{mmlu}
Dan Hendrycks, Collin Burns, Steven Basart, Andy Zou, Mantas Mazeika, Dawn Song, and Jacob Steinhardt.
\newblock Measuring massive multitask language understanding.
\newblock In \emph{International Conference on Learning Representations}, 2021.

\bibitem[Hu et~al.(2023)Hu, Zhou, Hergul, Gritta, Zhang, Iacobacci, Vuli{\'c}, and Korhonen]{multi3woz}
Songbo Hu, Han Zhou, Mete Hergul, Milan Gritta, Guchun Zhang, Ignacio Iacobacci, Ivan Vuli{\'c}, and Anna Korhonen.
\newblock Multi 3 woz: A multilingual, multi-domain, multi-parallel dataset for training and evaluating culturally adapted task-oriented dialog systems.
\newblock \emph{Transactions of the Association for Computational Linguistics}, 11:\penalty0 1396--1415, 2023.

\bibitem[Huang et~al.(2024)Huang, La~Malfa, Marro, Asperti, Cohn, and Wooldridge]{huang-etal-2024-notion}
X.~Angelo Huang, Emanuele La~Malfa, Samuele Marro, Andrea Asperti, Anthony~G. Cohn, and Michael~J. Wooldridge.
\newblock A notion of complexity for theory of mind via discrete world models.
\newblock In Yaser Al-Onaizan, Mohit Bansal, and Yun-Nung Chen, editors, \emph{Findings of the Association for Computational Linguistics: EMNLP 2024}, pages 2964--2983, Miami, Florida, USA, November 2024. Association for Computational Linguistics.
\newblock \doi{10.18653/v1/2024.findings-emnlp.167}.
\newblock \url{https://aclanthology.org/2024.findings-emnlp.167}.

\bibitem[Jiang et~al.(2024)Jiang, Sablayrolles, Roux, Mensch, Savary, Bamford, Chaplot, Casas, Hanna, Bressand, et~al.]{mixtral}
Albert~Q Jiang, Alexandre Sablayrolles, Antoine Roux, Arthur Mensch, Blanche Savary, Chris Bamford, Devendra~Singh Chaplot, Diego de~las Casas, Emma~Bou Hanna, Florian Bressand, et~al.
\newblock Mixtral of experts.
\newblock \emph{arXiv preprint arXiv:2401.04088}, 2024.

\bibitem[Jin et~al.(2024)Jin, Wu, Cao, Xiang, Kuo, Hu, Ullman, Torralba, Tenenbaum, and Shu]{mmtomqa}
Chuanyang Jin, Yutong Wu, Jing Cao, Jiannan Xiang, Yen-Ling Kuo, Zhiting Hu, Tomer Ullman, Antonio Torralba, Joshua Tenenbaum, and Tianmin Shu.
\newblock {MMT}o{M}-{QA}: Multimodal theory of mind question answering.
\newblock In Lun-Wei Ku, Andre Martins, and Vivek Srikumar, editors, \emph{Proceedings of the 62nd Annual Meeting of the Association for Computational Linguistics (Volume 1: Long Papers)}, pages 16077--16102, Bangkok, Thailand, August 2024. Association for Computational Linguistics.
\newblock \url{https://aclanthology.org/2024.acl-long.851}.

\bibitem[Jung et~al.(2024)Jung, Kim, Jin, Kim, Seonwoo, Choi, Oh, and Kim]{jung2024perceptions}
Chani Jung, Dongkwan Kim, Jiho Jin, Jiseon Kim, Yeon Seonwoo, Yejin Choi, Alice Oh, and Hyunwoo Kim.
\newblock Perceptions to beliefs: Exploring precursory inferences for theory of mind in large language models.
\newblock \emph{arXiv preprint arXiv:2407.06004}, 2024.

\bibitem[Kim et~al.(2023)Kim, Sclar, Zhou, Bras, Kim, Choi, and Sap]{fantom}
Hyunwoo Kim, Melanie Sclar, Xuhui Zhou, Ronan Bras, Gunhee Kim, Yejin Choi, and Maarten Sap.
\newblock Fantom: A benchmark for stress-testing machine theory of mind in interactions.
\newblock In \emph{Proceedings of the 2023 Conference on Empirical Methods in Natural Language Processing}, pages 14397--14413, 2023.

\bibitem[Kim et~al.(2024)Kim, Suk, Cho, Longpre, Kim, Yoon, Son, Cho, Shafayat, Baek, et~al.]{biggenbench}
Seungone Kim, Juyoung Suk, Ji~Yong Cho, Shayne Longpre, Chaeeun Kim, Dongkeun Yoon, Guijin Son, Yejin Cho, Sheikh Shafayat, Jinheon Baek, et~al.
\newblock The biggen bench: A principled benchmark for fine-grained evaluation of language models with language models.
\newblock \emph{arXiv preprint arXiv:2406.05761}, 2024.

\bibitem[Kinderman et~al.(1998)Kinderman, Dunbar, and Bentall]{kinderman1998theory}
Peter Kinderman, Robin Dunbar, and Richard~P Bentall.
\newblock Theory-of-mind deficits and causal attributions.
\newblock \emph{British journal of Psychology}, 89\penalty0 (2):\penalty0 191--204, 1998.

\bibitem[Le et~al.(2019)Le, Boureau, and Nickel]{le2019revisiting}
Matthew Le, Y-Lan Boureau, and Maximilian Nickel.
\newblock Revisiting the evaluation of theory of mind through question answering.
\newblock In \emph{Proceedings of the 2019 Conference on Empirical Methods in Natural Language Processing and the 9th International Joint Conference on Natural Language Processing (EMNLP-IJCNLP)}, pages 5872--5877, 2019.

\bibitem[Li et~al.(2024{\natexlab{a}})Li, Cui, Zhao, Kong, and Bi]{li2024gsm}
Qintong Li, Leyang Cui, Xueliang Zhao, Lingpeng Kong, and Wei Bi.
\newblock Gsm-plus: A comprehensive benchmark for evaluating the robustness of llms as mathematical problem solvers.
\newblock \emph{arXiv preprint arXiv:2402.19255}, 2024{\natexlab{a}}.

\bibitem[Li et~al.(2024{\natexlab{b}})Li, Liu, Liang, and Hashimoto]{autobencher}
Xiang~Lisa Li, Evan~Zheran Liu, Percy Liang, and Tatsunori Hashimoto.
\newblock Autobencher: Creating salient, novel, difficult datasets for language models, 2024{\natexlab{b}}.
\newblock \url{https://arxiv.org/abs/2407.08351}.

\bibitem[Luo et~al.(2024)Luo, Xu, Zhao, Sun, Geng, Hu, Tao, Ma, Lin, and Jiang]{luo2023wizardcoder}
Ziyang Luo, Can Xu, Pu~Zhao, Qingfeng Sun, Xiubo Geng, Wenxiang Hu, Chongyang Tao, Jing Ma, Qingwei Lin, and Daxin Jiang.
\newblock Wizardcoder: Empowering code large language models with evol-instruct.
\newblock 2024.
\newblock \url{https://openreview.net/forum?id=UnUwSIgK5W}.

\bibitem[Lupidi et~al.(2024)Lupidi, Gemmell, Cancedda, Dwivedi-Yu, Weston, Foerster, Raileanu, and Lomeli]{lupidi2024source2synth}
Alisia Lupidi, Carlos Gemmell, Nicola Cancedda, Jane Dwivedi-Yu, Jason Weston, Jakob Foerster, Roberta Raileanu, and Maria Lomeli.
\newblock Source2synth: Synthetic data generation and curation grounded in real data sources.
\newblock \emph{arXiv preprint arXiv:2409.08239}, 2024.

\bibitem[OpenAI(2024)]{gpt4o}
OpenAI, 2024.
\newblock \url{https://openai.com/index/hello-gpt-4o}.

\bibitem[Premack and Woodruff(1978)]{premack1978does}
David Premack and Guy Woodruff.
\newblock Does the chimpanzee have a theory of mind?
\newblock \emph{Behavioral and brain sciences}, 1\penalty0 (4):\penalty0 515--526, 1978.

\bibitem[Rabinowitz et~al.(2018)Rabinowitz, Perbet, Song, Zhang, Eslami, and Botvinick]{rabinowitz2018machine}
Neil Rabinowitz, Frank Perbet, Francis Song, Chiyuan Zhang, SM~Ali Eslami, and Matthew Botvinick.
\newblock Machine theory of mind.
\newblock In \emph{International conference on machine learning}, pages 4218--4227. PMLR, 2018.

\bibitem[Sap et~al.(2022)Sap, LeBras, Fried, and Choi]{sap2022neural}
Maarten Sap, Ronan LeBras, Daniel Fried, and Yejin Choi.
\newblock Neural theory-of-mind? on the limits of social intelligence in large lms.
\newblock In \emph{Proceedings of the Association for Computational Linguistics: EMNLP 2022}, page 3762–3780, Abu Dhabi, United Arab Emirates, December 2022. Association for Computational Linguistics.
\newblock \url{https://preview.aclanthology.org/emnlp-22-ingestion/2022.emnlp-main.248}.

\bibitem[Sclar et~al.(2022)Sclar, Neubig, and Bisk]{sclar22symmetric}
Melanie Sclar, Graham Neubig, and Yonatan Bisk.
\newblock Symmetric machine theory of mind.
\newblock In Kamalika Chaudhuri, Stefanie Jegelka, Le~Song, Csaba Szepesvari, Gang Niu, and Sivan Sabato, editors, \emph{Proceedings of the 39th International Conference on Machine Learning}, volume 162 of \emph{Proceedings of Machine Learning Research}, pages 19450--19466. PMLR, 17--23 Jul 2022.
\newblock \url{https://proceedings.mlr.press/v162/sclar22a.html}.

\bibitem[Sclar et~al.(2023)Sclar, Kumar, West, Suhr, Choi, and Tsvetkov]{symbolictom}
Melanie Sclar, Sachin Kumar, Peter West, Alane Suhr, Yejin Choi, and Yulia Tsvetkov.
\newblock Minding language models{'} (lack of) theory of mind: A plug-and-play multi-character belief tracker.
\newblock In Anna Rogers, Jordan Boyd-Graber, and Naoaki Okazaki, editors, \emph{Proceedings of the 61st Annual Meeting of the Association for Computational Linguistics (Volume 1: Long Papers)}, pages 13960--13980, Toronto, Canada, July 2023. Association for Computational Linguistics.
\newblock \doi{10.18653/v1/2023.acl-long.780}.
\newblock \url{https://aclanthology.org/2023.acl-long.780}.

\bibitem[Shamay-Tsoory et~al.(2010)Shamay-Tsoory, Harari, Aharon-Peretz, and Levkovitz]{shamay2010role}
Simone~G Shamay-Tsoory, Hagai Harari, Judith Aharon-Peretz, and Yechiel Levkovitz.
\newblock The role of the orbitofrontal cortex in affective theory of mind deficits in criminal offenders with psychopathic tendencies.
\newblock \emph{Cortex}, 46\penalty0 (5):\penalty0 668--677, 2010.

\bibitem[Shapira et~al.(2023{\natexlab{a}})Shapira, Levy, Alavi, Zhou, Choi, Goldberg, Sap, and Shwartz]{shapira2023clever}
Natalie Shapira, Mosh Levy, Seyed~Hossein Alavi, Xuhui Zhou, Yejin Choi, Yoav Goldberg, Maarten Sap, and Vered Shwartz.
\newblock Clever hans or neural theory of mind? stress testing social reasoning in large language models, 2023{\natexlab{a}}.

\bibitem[Shapira et~al.(2023{\natexlab{b}})Shapira, Zwirn, and Goldberg]{fauxpas}
Natalie Shapira, Guy Zwirn, and Yoav Goldberg.
\newblock How well do large language models perform on faux pas tests?
\newblock In \emph{Findings of the Association for Computational Linguistics: ACL 2023}, pages 10438--10451, 2023{\natexlab{b}}.

\bibitem[Shi et~al.(2024)Shi, Ye, Fang, Jin, Isik, Kuo, and Shu]{shi2024muma}
Haojun Shi, Suyu Ye, Xinyu Fang, Chuanyang Jin, Layla Isik, Yen-Ling Kuo, and Tianmin Shu.
\newblock Muma-tom: Multi-modal multi-agent theory of mind.
\newblock \emph{arXiv preprint arXiv:2408.12574}, 2024.

\bibitem[Strachan et~al.(2024)Strachan, Albergo, Borghini, Pansardi, Scaliti, Gupta, Saxena, Rufo, Panzeri, Manzi, et~al.]{strachan2024naturepaper}
James~WA Strachan, Dalila Albergo, Giulia Borghini, Oriana Pansardi, Eugenio Scaliti, Saurabh Gupta, Krati Saxena, Alessandro Rufo, Stefano Panzeri, Guido Manzi, et~al.
\newblock Testing theory of mind in large language models and humans.
\newblock \emph{Nature Human Behaviour}, pages 1--11, 2024.

\bibitem[Ullman(2023)]{ullman2023large}
Tomer Ullman.
\newblock Large language models fail on trivial alterations to theory-of-mind tasks.
\newblock \emph{arXiv preprint arXiv:2302.08399}, 2023.

\bibitem[Van~Breugel et~al.(2023)Van~Breugel, Qian, and Van Der~Schaar]{van2023synthetic}
Boris Van~Breugel, Zhaozhi Qian, and Mihaela Van Der~Schaar.
\newblock Synthetic data, real errors: how (not) to publish and use synthetic data.
\newblock In \emph{International Conference on Machine Learning}, pages 34793--34808. PMLR, 2023.

\bibitem[Wang et~al.(2021)Wang, Saha, Gregori, Joyner, and Goel]{wang2021hcitom}
Qiaosi Wang, Koustuv Saha, Eric Gregori, David Joyner, and Ashok Goel.
\newblock Towards mutual theory of mind in human-ai interaction: How language reflects what students perceive about a virtual teaching assistant.
\newblock In \emph{Proceedings of the 2021 CHI Conference on Human Factors in Computing Systems}, pages 1--14, 2021.

\bibitem[Wang et~al.(2024)Wang, Kulikov, Golovneva, Yu, Yuan, Dwivedi-Yu, Pang, Fazel-Zarandi, Weston, and Li]{wang2024self}
Tianlu Wang, Ilia Kulikov, Olga Golovneva, Ping Yu, Weizhe Yuan, Jane Dwivedi-Yu, Richard~Yuanzhe Pang, Maryam Fazel-Zarandi, Jason Weston, and Xian Li.
\newblock Self-taught evaluators.
\newblock \emph{arXiv preprint arXiv:2408.02666}, 2024.

\bibitem[Wang et~al.(2023)Wang, Kordi, Mishra, Liu, Smith, Khashabi, and Hajishirzi]{wang-etal-2023-self-instruct}
Yizhong Wang, Yeganeh Kordi, Swaroop Mishra, Alisa Liu, Noah~A. Smith, Daniel Khashabi, and Hannaneh Hajishirzi.
\newblock Self-instruct: Aligning language models with self-generated instructions.
\newblock In Anna Rogers, Jordan Boyd-Graber, and Naoaki Okazaki, editors, \emph{Proceedings of the 61st Annual Meeting of the Association for Computational Linguistics (Volume 1: Long Papers)}, pages 13484--13508, Toronto, Canada, July 2023. Association for Computational Linguistics.
\newblock \doi{10.18653/v1/2023.acl-long.754}.
\newblock \url{https://aclanthology.org/2023.acl-long.754}.

\bibitem[Wellman(2014)]{wellman2014making}
Henry~M Wellman.
\newblock \emph{Making minds: How theory of mind develops}.
\newblock Oxford University Press, 2014.

\bibitem[West et~al.(2022)West, Bhagavatula, Hessel, Hwang, Jiang, Le~Bras, Lu, Welleck, and Choi]{west-etal-2022-symbolic}
Peter West, Chandra Bhagavatula, Jack Hessel, Jena Hwang, Liwei Jiang, Ronan Le~Bras, Ximing Lu, Sean Welleck, and Yejin Choi.
\newblock Symbolic knowledge distillation: from general language models to commonsense models.
\newblock In Marine Carpuat, Marie-Catherine de~Marneffe, and Ivan~Vladimir Meza~Ruiz, editors, \emph{Proceedings of the 2022 Conference of the North American Chapter of the Association for Computational Linguistics: Human Language Technologies}, pages 4602--4625, Seattle, United States, July 2022. Association for Computational Linguistics.
\newblock \doi{10.18653/v1/2022.naacl-main.341}.
\newblock \url{https://aclanthology.org/2022.naacl-main.341}.

\bibitem[Wilf et~al.(2023)Wilf, Lee, Liang, and Morency]{simtom}
Alex Wilf, Sihyun~Shawn Lee, Paul~Pu Liang, and Louis-Philippe Morency.
\newblock Think twice: Perspective-taking improves large language models' theory-of-mind capabilities.
\newblock \emph{arXiv preprint arXiv:2311.10227}, 2023.

\bibitem[Wimmer and Perner(1983)]{wimmer1983beliefs}
Heinz Wimmer and Josef Perner.
\newblock Beliefs about beliefs: Representation and constraining function of wrong beliefs in young children's understanding of deception.
\newblock \emph{Cognition}, 13\penalty0 (1):\penalty0 103--128, 1983.

\bibitem[Wood et~al.(2021)Wood, Baltru{\v{s}}aitis, Hewitt, Dziadzio, Cashman, and Shotton]{wood2021fake}
Erroll Wood, Tadas Baltru{\v{s}}aitis, Charlie Hewitt, Sebastian Dziadzio, Thomas~J Cashman, and Jamie Shotton.
\newblock Fake it till you make it: face analysis in the wild using synthetic data alone.
\newblock In \emph{Proceedings of the IEEE/CVF international conference on computer vision}, pages 3681--3691, 2021.

\bibitem[Wu et~al.(2023)Wu, He, Jia, Mihalcea, Chen, and Deng]{hitom}
Yufan Wu, Yinghui He, Yilin Jia, Rada Mihalcea, Yulong Chen, and Naihao Deng.
\newblock Hi-{T}o{M}: A benchmark for evaluating higher-order theory of mind reasoning in large language models.
\newblock pages 10691--10706, December 2023.
\newblock \doi{10.18653/v1/2023.findings-emnlp.717}.
\newblock \url{https://aclanthology.org/2023.findings-emnlp.717}.

\bibitem[Xu et~al.(2024)Xu, Zhao, Zhu, Du, and He]{opentom}
Hainiu Xu, Runcong Zhao, Lixing Zhu, Jinhua Du, and Yulan He.
\newblock Opentom: A comprehensive benchmark for evaluating theory-of-mind reasoning capabilities of large language models.
\newblock \emph{arXiv preprint arXiv:2402.06044}, 2024.

\bibitem[Yang et~al.(2022)Yang, Tian, Peng, and Klein]{yang2022re3}
Kevin Yang, Yuandong Tian, Nanyun Peng, and Dan Klein.
\newblock Re3: Generating longer stories with recursive reprompting and revision.
\newblock pages 4393--4479, December 2022.
\newblock \doi{10.18653/v1/2022.emnlp-main.296}.
\newblock \url{https://aclanthology.org/2022.emnlp-main.296}.

\bibitem[Yu et~al.(2024)Yu, Jiang, Shi, YU, Liu, Zhang, Kwok, Li, Weller, and Liu]{yu2023metamath}
Longhui Yu, Weisen Jiang, Han Shi, Jincheng YU, Zhengying Liu, Yu~Zhang, James Kwok, Zhenguo Li, Adrian Weller, and Weiyang Liu.
\newblock Metamath: Bootstrap your own mathematical questions for large language models.
\newblock 2024.
\newblock \url{https://openreview.net/forum?id=N8N0hgNDRt}.

\bibitem[Zelikman et~al.(2022)Zelikman, Wu, Mu, and Goodman]{zelikman2022star}
Eric Zelikman, Yuhuai Wu, Jesse Mu, and Noah Goodman.
\newblock Star: Bootstrapping reasoning with reasoning.
\newblock \emph{Advances in Neural Information Processing Systems}, 35:\penalty0 15476--15488, 2022.

\bibitem[Zhang et~al.(2023)Zhang, Li, Cui, Cai, Liu, Fu, Huang, Zhao, Zhang, Chen, et~al.]{zhang2023siren}
Yue Zhang, Yafu Li, Leyang Cui, Deng Cai, Lemao Liu, Tingchen Fu, Xinting Huang, Enbo Zhao, Yu~Zhang, Yulong Chen, et~al.
\newblock Siren's song in the ai ocean: a survey on hallucination in large language models.
\newblock \emph{arXiv preprint arXiv:2309.01219}, 2023.

\bibitem[Zhou et~al.(2023)Zhou, Madaan, Potharaju, Gupta, McKee, Holtzman, Pujara, Ren, Mishra, Nematzadeh, et~al.]{t4d}
Pei Zhou, Aman Madaan, Srividya~Pranavi Potharaju, Aditya Gupta, Kevin~R McKee, Ari Holtzman, Jay Pujara, Xiang Ren, Swaroop Mishra, Aida Nematzadeh, et~al.
\newblock How far are large language models from agents with theory-of-mind?
\newblock \emph{arXiv preprint arXiv:2310.03051}, 2023.

\bibitem[Zhu et~al.(2021)Zhu, Neubig, and Bisk]{zhu2021few}
Hao Zhu, Graham Neubig, and Yonatan Bisk.
\newblock Few-shot language coordination by modeling theory of mind.
\newblock In \emph{International Conference on Machine Learning}, pages 12901--12911. PMLR, 2021.

\end{thebibliography}

\clearpage
\newpage
\beginappendix
\appendix
\section{Appendix}

\subsection{Actions' formal definition (cont. from \ref{sec:methods_dsl})}\label{app:formal}
All actions are functions that transform a state into another state, updating the world state and the beliefs of everyone involved up to two levels of recursion. All actions have preconditions, e.g. to enter a room you need to not be in it already.

A state $\in\mathcal{S}$ is comprised of a world state $ws$ (the things currently true physically about the world described), the first-order beliefs $b_1$, and the second-order beliefs $b_2$. First-order beliefs describe what each person believes to be the current world state, e.g. Anne believes that the apple is salted. Second-order beliefs describe what each person estimates that each other person believes to be the current world state, e.g. Anne believes that Beth thinks that the apple is salted.

Let's describe the definition of leaving in a room through an example: ``Beth left the kitchen.'', and build the definition of the action function $a_{\text{leave, Beth, kitchen}}: \mathcal{S} \rightarrow \mathcal{S}$. As described above, the state is comprised of a world state, first-order beliefs, and second-order beliefs, i.e.,
\begin{align*}
    a_{\text{leave, Beth, kitchen}}(ws, b_1, b_2) := (ws', b_1', b_2')
\end{align*}

Let's first describe the world state update $ws'$. The world state remains the same for every entity (object, container, person, etc.), except for the person leaving the room---Beth. Thus, $$ws(q, \text{room})~=~ws'(q, \text{room}) \ \  \forall q \neq \text{Beth} \text{ \ \ \ and \ \ \ } ws'(q, \text{room}) = \neg \text{kitchen}$$

Let's then describe the first-order belief updates $b_1'$. Here, we assume that everyone in the same room as Beth (the kitchen) will know that Beth has left. We denote this group of people as $\text{witnesses(kitchen)}$: $$\text{witnesses(kitchen)}:=\{p | ws(p, \text{room}) = \text{kitchen}\}$$
Everyone not in the kitchen will assume that Beth is still there unless communicated otherwise, since they have no reason to believe she has left. Thus, $$b_1(p, \text{Beth}, \text{room}) = b_1'(p, \text{Beth}, \text{room}) = \text{kitchen} \ \ \forall p \notin \text{witnesses(kitchen)}$$
$$b_1(p, \text{Beth}, \text{room}) = \neg \text{kitchen} \ \ \forall p \in \text{witnesses(kitchen)}$$

We now describe the second-order belief updates $b_2'$. Here, we assume that everyone in the kitchen (including Beth) assumes that everyone else in the kitchen knows Beth left (and only them). If someone was not in the kitchen, they will assume nothing has happened. Formally,
$$b_2(p, q, \text{Beth}, \text{room}) = b_2'(p, q, \text{Beth}, \text{room}) = \text{kitchen} \ \ \ \forall p \notin \text{witnesses(kitchen)}, \ \ \forall q$$
$$b_2(p, q, \text{Beth}, \text{room}) = \neg \text{kitchen} \ \ \forall p \in \text{witnesses(kitchen)} \ \ \forall q \in \text{witnesses(kitchen)}$$
$$b_2(p, q, \text{Beth}, \text{room}) = \text{kitchen} \ \ \forall p \in \text{witnesses(kitchen)} \ \ \forall q \notin \text{witnesses(kitchen)}$$

Finally, the function can only be applied if Beth is in the kitchen, i.e. it has the precondition $ws(\text{Beth}, \text{room}) = \text{kitchen}$.

All other functions definitions can be found verbatim in the code to be released.

\newpage
\subsection{All supported questions (Cont. from Section~\ref{sec:methods_question})}\label{app:all_questions}

\begin{table}[h]
\centering
\caption{List of all supported \methodname{} questions per property discussed and level of theory of mind, transcribed verbatim.}
\begin{tabular}{@{}p{1in}p{0.3in}p{2.6in}p{1in}@{}}
\toprule
\begin{tabular}[c]{@{}l@{}}Property \\ asked about\end{tabular} & \begin{tabular}[c]{@{}l@{}}ToM \\ Order\end{tabular} & Question (requesting \textit{Short Answer.} in prompt) & \begin{tabular}[c]{@{}l@{}}Expected \\ Answers\end{tabular} \\ \midrule \toprule
room location & -- & In which room was the $<$object$>$ at the beginning? & room name \\ \midrule
room location & -- & In which room is the $<$object$>$ now? & room name \\ \midrule
room location & -- & In which room was the $<$object$>$ before $<$action$>$? & room name \\ \midrule
room location & 1st & In which room will $<$person$>$ search for the $<$object$>$? & room name \\ \midrule
room location & 2nd & In which room does $<$person1$>$ think that $<$person2$>$ will search for the $<$object$>$? & room name \\ \midrule
container location & -- & In which container was the $<$object$>$ at the beginning? & container name \\ \midrule
container location & -- & In which container is the $<$object$>$ now? & container name \\ \midrule
container location & -- & In which container was the $<$object$>$ before $<$action$>$? & container name \\ \midrule
container location & 1st & In which container will $<$person$>$ search for the $<$object$>$? & container name \\ \midrule
container location & 2nd & In which container does $<$person1$>$ think that $<$person2$>$ will search for the $<$object$>$? & container name \\ \midrule
abstract topic knowledge & 1st & Does $<$person1$>$ know about $<$topicDiscussed$>$? & \textit{yes} or \textit{no}  \\ \midrule
abstract topic knowledge & 2nd & What does $<$person1$>$ think about $<$person2$>$'s belief on $<$topicDiscussed$>$? (knows about it / does not know about it) & \textit{knows about it} or \textit{does not know about it}  \\ \midrule
knowledge about state update & 1st & Does $<$person$>$ believe that the $<$object$>$ $<$newState$>$? Answer yes or no. & \textit{yes} or \textit{no}  \\ \midrule
knowledge about state update & 2nd & Does $<$person1$>$ believe that $<$person2$>$ believes that the $<$object$>$ $<$newState$>$? Answer yes or no. & \textit{yes} or \textit{no}  \\ 
\bottomrule
\end{tabular}
\end{table}

\newpage
\section{Additional Experiments}

\subsection{A*-generated stories are more challenging than overgenerating and filtering (cont. from $\S$~\ref{sec:ttm_eval})}
\begin{figure}[h]
    \centering
    \includegraphics[width=0.6\linewidth]{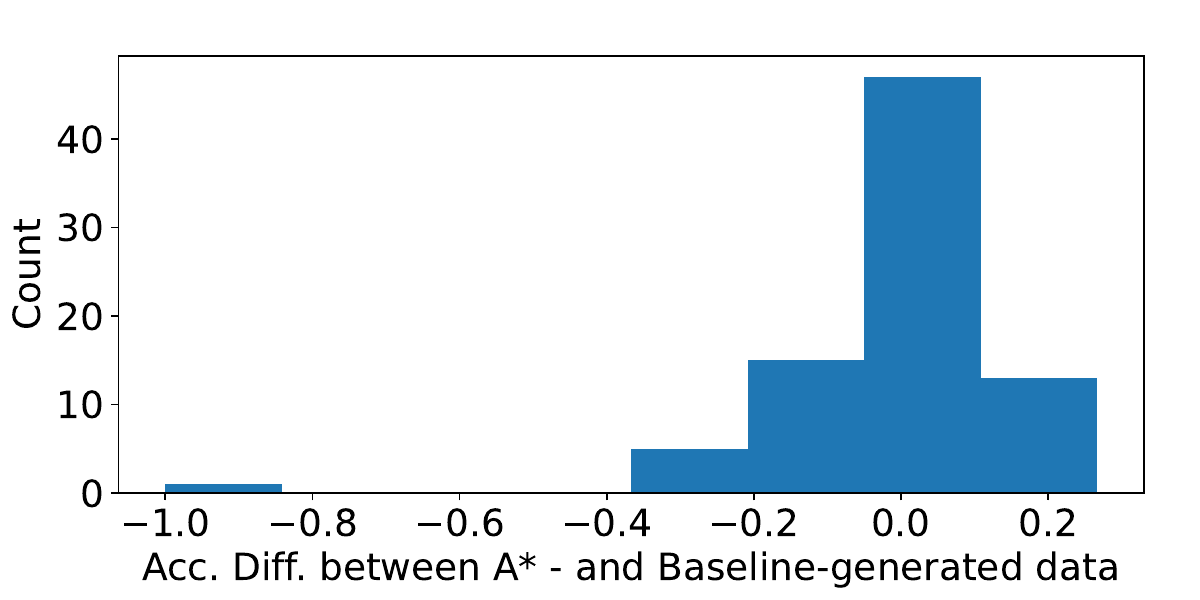}  
    \caption{Histogram depicting accuracy differences between A*-generated datasets for Llama-3.1-70B-Instruct and a dataset created by over-generating and filtering with the same budget (i.e., baseline). Results show that A* is better at finding story structures that make a challenging benchmark by showing low accuracy (negative values mean A* is better at finding challenging story structures).
}
    \label{app:astar_vs_baseline}
\end{figure}

\subsection{Infilled Story Structures Remain Challenging (cont. from $\S$~\ref{sec:ttm_eval})}\label{app:infilling}
\begin{table}[h!]
\centering
\caption{Changes in accuracy when infilling \methodname{}-generated story structures to output natural-sounding stories. We only include comparison between the 40\% of stories where the LLM as a judge (Llama-3.1-70B Instruct) determined that all infilled actions were high quality.}\label{tab:infilling}
\begin{tabular}{cccccc}
\toprule
\begin{tabular}[c]{@{}c@{}}Action Set:\\$\{a_{\text{enter}}, a_{\text{leave}}, \ldots$\end{tabular} & \begin{tabular}[c]{@{}c@{}}Include\\ asymmetry\end{tabular} & \begin{tabular}[c]{@{}c@{}}Acc. Story\\Structure\end{tabular} & \begin{tabular}[c]{@{}c@{}}Acc.\\Infilled\end{tabular} & \begin{tabular}[c]{@{}c@{}}Acc.\\Diff.\end{tabular} \\ \midrule \toprule
\multirow{2}{*}{$\ldots, {a_{\text{moveObjContainer}}}\}$, \ \ \ ($\mathbf{\text{denoted }\mathcal{A}_1}$)} & \xmark & 0.20 & 0.43 & 0.22 \\
 & \cmark & 0.02 & 0.36 & 0.35 \\ \midrule
\multirow{2}{*}{$\ldots, {a_{\text{updateObjState}}}\}$, \ \ \ ($\mathbf{\text{denoted }\mathcal{A}_2}$)} & \xmark & 0.40 & 0.44 & 0.03 \\
  & \cmark  & 0.49 & 0.48 & -0.01 \\ \midrule
\multirow{2}{*}{$\ldots, {a_{\text{moveObjContainer}}}, {a_{\text{updateObjState}}}\}$ \ \ ($\mathbf{\mathcal{A}_3}$)  } & \xmark & 0.45 & 0.48 & 0.03 \\
  & \cmark  & 0.17 & 0.60 & 0.43 \\ \midrule
\multirow{2}{*}{$\ldots, a_{\text{moveObjContainer}}, {a_{\text{moveObjRoom}}}\}$  \ \ ($\mathbf{\mathcal{A}_4}$) }  & \xmark & 0.12 & 0.73 & 0.62 \\
  & \cmark  & 0.13 & 0.37 & 0.23 \\ \midrule
\multirow{2}{*}{$\ldots, a_{\text{moveObjContainer}}, {a_{\text{info-*}}}\}$  \ \ ($\mathbf{\mathcal{A}_5}$)  }  & \xmark & 0.09 & 0.45 & 0.37 \\
  & \cmark  & 0.10 & 0.42 & 0.32 \\ \midrule
\multirow{2}{*}{{\small $\ldots, a_{\text{moveObjContainer}}, a_{\text{moveObjRoom}}, {a_{\text{info-*}}} \}$}  \ \ ($\mathbf{\mathcal{A}_6}$)  } & \xmark & 0.08 & 0.31 & 0.23 \\
 & \cmark  & 0.13 & 0.51 & 0.38 \\ \midrule
\multirow{2}{*}{{\small $a_{\text{moveObjContainer}}, a_{\text{moveObjRoom}}, a_{\text{chitChat-*}}, {a_{\text{info-*}}} \}$} ($\mathbf{\mathcal{A}_7}$) } & \xmark & 0.73 & 0.86 & 0.13 \\
 & \cmark  & 0.75 & 0.85 & 0.10 \\ \midrule
\multirow{2}{*}{$\ldots, {a_{\text{chitChat-private}}} \}$ \ \ ($\mathbf{\mathcal{A}_8}$) } & \xmark & 0.75 & 0.82 & 0.07 \\
  & \cmark & 0.68 & 0.76 & 0.08 \\ \midrule
\multirow{2}{*}{$\ldots, {a_{\text{chitChat-public}}} \}$ \ \ ($\mathbf{\mathcal{A}_9}$) }  & \xmark & 0.61 & 0.68 & 0.07 \\
 & \cmark & 0.54 & 0.61 & 0.07 \\
\bottomrule
\end{tabular}
\end{table}

Table~\ref{tab:infilling} shows a breakdown across all settings. 40\% of the stories were infilled in a single attempt. Next step options are sampled simultaneously with repetition penalty for added wording diversity. This is due to the stringent LLM as a judge conditions to ensure quality. Without these quality and diversity constraints, 81\% of the story structures are infilled within a single attempt (73\% with only quality constraints).

\subsection{Models Fail Both at Theory of Mind and Pure State Tracking (Cont. from $\S$~\ref{skills_of_tom})}\label{app:int_vs_no_int}


\begin{table}[h!]
\centering\caption{Accuracy breakdown of the experiment shown in Table~\ref{tab:eval_benchmark_accuracy}, discriminating if each question is \textit{interesting} or not. A question is interesting if the answer would change depending on the entity asked about, thus potentially requiring theory of mind. Results show that part of a model's difficulty with \methodname{}'s generated data could be attributed to poor state tracking (i.e., the uninteresting questions, noted \textlnot Int.).}\label{tab:int_vs_no_int}
\begin{tabular}{@{}ccccccccccc@{}}
\toprule
 & & \multicolumn{3}{c}{Llama}   & \multicolumn{3}{c}{GPT4o}   & \multicolumn{3}{c}{Mixtral}   \\ \midrule
Action Set & \begin{tabular}[c]{@{}c@{}}Includes\\ Symmetry?\end{tabular} & \begin{tabular}[c]{@{}c@{}}Acc.\\ Int.\end{tabular} & \begin{tabular}[c]{@{}c@{}}Acc.\\ \textlnot Int.\end{tabular} & \begin{tabular}[c]{@{}c@{}}\%\\ Int.\end{tabular} & \begin{tabular}[c]{@{}c@{}}Acc.\\ Int.\end{tabular} & \begin{tabular}[c]{@{}c@{}}Acc.\\ \textlnot Int.\end{tabular} & \begin{tabular}[c]{@{}c@{}}\%\\ Int.\end{tabular}  & \begin{tabular}[c]{@{}c@{}}Acc.\\ Int.\end{tabular} & \begin{tabular}[c]{@{}c@{}}Acc.\\ \textlnot Int.\end{tabular} & \begin{tabular}[c]{@{}c@{}}\%\\ Int.\end{tabular}  \\ \midrule \toprule

$\mathbf{\mathcal{A}_1}$ & \xmark &0.42 & 0.20 & 44\% & 0.47 & 0.43 & 45\% & 0.44 & 0.38 & 48\%  \\
 & \cmark & 0.31 & 0.01 & 49\% &  0.46 & 0.28 & 49\%   & 0.37 & 0.40 & 50\% \\ \midrule
$\mathbf{\mathcal{A}_2}$ & \xmark & 0.61 & 0.25 & 42\%  &  0.57 & 0.24 & 41\%  & 0.39 & 0.01 & 50\% \\
 & \cmark & 0.49 & 0.22 & 50\% &  0.56 & 0.23 & 49\% & 0.19 & 0.00 & 34\% \\ \midrule
$\mathbf{\mathcal{A}_3}$ & \xmark & 0.54 & 0.24 & 47\% & 0.55 & 0.35 & 47\%   & 0.46 & 0.18 & 46\% \\
 & \cmark & 0.42 & 0.04 & 50\% &  0.57 & 0.31 & 48\%  & 0.36 & 0.04 & 48\% \\ \midrule
$\mathbf{\mathcal{A}_4}$ & \xmark & 0.56 & 0.11 & 16\% & 0.58 & 0.09 & 25\%  & 0.49 & 0.00 & 25\% \\
 & \cmark & 0.25 & 0.18 & 49\% &  0.47 & 0.21 & 35\%  & 0.44 & 0.00 & 31\% \\ \midrule
$\mathbf{\mathcal{A}_5}$ & \xmark & 0.36 & 0.09 & 50\% & 0.50 & 0.30 & 45\% & 0.44 & 0.40 & 48\% \\
 & \cmark & 0.24 & 0.15 & 50\% & 0.44 & 0.29 & 48\% & 0.44 & 0.47 & 50\% \\ \midrule
$\mathbf{\mathcal{A}_6}$ & \xmark &  0.37 & 0.18 & 50\% & 0.64 & 0.25 & 44\%   & 0.50 & 0.03 & 48\% \\
 & \cmark & 0.31 & 0.14 & 50\% & 0.64 & 0.25 & 41\%  & 0.49 & 0.04 & 48\% \\ \midrule
$\mathbf{\mathcal{A}_7}$ & \xmark & 0.74 & 0.74 & 46\% & 0.76 & 0.76 & 44\% & 0.58 & 0.72 & 42\% \\
 & \cmark & 0.74 & 0.67 & 43\% & 0.72 & 0.75 & 47\% & 0.47 & 0.67 & 42\% \\ \midrule
$\mathbf{\mathcal{A}_8}$ & \xmark & 0.81 & 0.62 & 67\% & 0.80 & 0.72 & 66\% & 0.55 & 0.42 & 67\% \\
 & \cmark & 0.62 & 0.59 & 74\% &  0.63 & 0.37 & 75\%  & 0.41 & 0.42 & 74\% \\ \midrule
$\mathbf{\mathcal{A}_9}$ & \xmark & 0.48 & 0.67 & 40\% & 0.50 & 0.42 & 40\%  & 0.52 & 0.27 & 40\% \\
 & \cmark & 0.54 & 0.51 & 50\% &  0.65 & 0.41 & 50\% &  0.54 & 0.27 & 50\% \\ \toprule
Total & --- & 0.49  & 0.31 & 48\% &  0.58 & 0.37 & 47\% & 0.45 & 0.26 & 47\% \\ 
\bottomrule 
\end{tabular}
\end{table}

\subsection{How likely is a randomly-sampled story to require theory of mind? (cont. from $\S$\ref{skills_of_tom})}\label{app:random_sample_stories}

\begin{table}[h!]
\centering
\caption{Probability that a randomly-sampled story would require theory of mind for answering at least one question. Actions considered are \{$a_{\text{enter}}$, $a_{\text{leave}}$, $a_{\text{moveObjContainer}}$\}, all settings of $\{2, 3, 4\}$ people and $\{2, 3, 4\}$ $a_{\text{moveObjContainer}}$ movements, with 10 maximum actions, are shown.}\label{tab:random_sample_stories}
\begin{tabular}{@{}cccc@{}}
\toprule
          & \multicolumn{3}{c}{Number of movements} \\ \midrule
Number of people & 2        & 3       & 4       \\ \toprule
2         & 0.131  &  0.208 & 0.235 \\
3         & 0.195   & 0.234 & 0.288  \\
4         &  0.210    & 0.259   & 0.315  \\ \bottomrule
\end{tabular}
\end{table}

\begin{table}[h!]
\centering
\caption{Probability that a randomly-sampled (story, question) pair would potentially require theory of mind, meaning that the answer to the question varies depending on the entities considered. Actions considered are \{$a_{\text{enter}}$, $a_{\text{leave}}$, $a_{\text{moveObjContainer}}$\}, all settings of $\{2, 3, 4\}$ people and $\{2, 3, 4\}$ $a_{\text{moveObjContainer}}$ movements, with 10 maximum actions, are shown.}\label{tab:random_sample_stories_interesting}
\begin{tabular}{@{}cccc@{}}
\toprule
          & \multicolumn{3}{c}{Number of movements} \\ \midrule
Number of people & 2        & 3       & 4       \\ \toprule
2         & 0.090  &  0.123 & 0.124 \\
3         & 0.120   & 0.109 & 0.121  \\
4         &  0.111    & 0.101   & 0.112  \\ \bottomrule
\end{tabular}
\end{table}

\begin{table}[h!]
\centering
\caption{Probability that a randomly-sampled (story, question, answer) triple would require an answer that is different from the true world state (i.e., it is a \textit{false-belief question}. Actions considered are \{$a_{\text{enter}}$, $a_{\text{leave}}$, $a_{\text{moveObjContainer}}$\}, all settings of $\{2, 3, 4\}$ people and $\{2, 3, 4\}$ $a_{\text{moveObjContainer}}$ movements, with 10 maximum actions, are shown.}\label{tab:random_sample_stories_fb}
\begin{tabular}{@{}cccc@{}}
\toprule
          & \multicolumn{3}{c}{Number of movements} \\ \midrule
Number of people & 2        & 3       & 4       \\ \toprule
2         & 0.059  &  0.084 & 0.086 \\
3         & 0.065   & 0.065 & 0.072  \\
4         &  0.056    & 0.049   & 0.058  \\ \bottomrule
\end{tabular}
\end{table}

\newpage
\subsection{On why a story with a greater number of people may counterintuitively imply a lower average difficulty}\label{app:counterintuitive_phenomenon}
In \methodname{}, the number of people and actions is important from a controllability and diversity perspective, but does not directly quantify task difficulty---difficulty quantification for theory of mind is an active area of research. \cite{huang-etal-2024-notion} quantifies a theory of mind problem complexity as the number of states necessary to solve it correctly (note that their approach requires manual annotation). While the number of states tends to increase with the number of people and actions, many questions do not require analyzing the whole story, e.g. if someone only entered the scene right at the end. When randomly sampling stories while fixing the number of core actions to e.g. 5, it’s more likely to have some characters with little involvement in the scene if there are 5 people in total than if there are 2 people. Since accuracy is computed across all questions about all characters, having a larger number of people may bump the average accuracy. \methodname{}'s flexible framework allows for minimizing these cases through modifying the lookahead, but we chose against both doing this or filtering questions to show the performance is low even without these considerations.

\newpage
\section{Prompts used for generating and validating \methodname{}'s data}\label{app:prompts_used}

\subsection{Generating story contexts (cont. from \S\ref{sec:storysampling})}
\begin{figure}[h]
    \centering
\begin{tcolorbox}[
        colback=promptboxcolor,    
        colframe=black,    
        boxrule=0.5pt,        
        arc=0mm,              
        left=1mm,             
        right=1mm,            
        top=1mm,              
        bottom=1mm]           
\lstset{
  basicstyle=\ttfamily,
  breaklines=true,
  breakatwhitespace=false,
  columns=fullflexible,
  keepspaces=true,
  frame=none,
}
\begin{lstlisting}[breakindent=0pt]
Suggest a short context where {num_people} people are together in a room. It should be at most two sentences long, and they should be able to observe each other. Later in the story, characters are going to move around and store objects, so your context should be plausible under those constraints. Do not explicitly include that they can all see each other, it should be clear from context. The room could be in a house, work environment, etc.

Here's an example for three people. Follow the same format.

LIST CHARACTERS' NAMES:
1. Emily, a meticulous office manager.
2. Jason, a tech-savvy intern.
3. Karen, a diligent accountant.

GIVE SHORT STORY CONTEXT:
Emily, Jason, and Karen gathered around the central table in the sleek office's conference room, discussing the upcoming audit. As they strategized, the shelves and storage compartments lining the walls around them held the tools and documents they would soon need to organize and pack away.

ROOM IN WHICH THIS STORY BEGINS:

NAME ONE REASONABLE ALTERNATIVE ROOM THEY COULD MOVE TO:

NAME ONE OBJECT TO BE MOVED BY A PERSON DURING THE STORY:

LIST {num_containers} REASONABLE OPAQUE CONTAINERS THAT COULD CONTAIN THIS OBJECT:

LIST {num_topics} DISTINCT AND REASONABLE TOPICS THEY COULD BE CHATTING ABOUT:

To get inspired, make this context happen in {sampled_location}. Suggested names are {sampled_names}, but feel free to come up with your own names if it would suit the story better. Be direct with your answers: do not include parentheses or clarifications beyond the responses requested. Do not refer to plural objects or give options if a singular thing is requested. The object could be anything--an apple, a pen, a spoon, a pair of scissors, a chocolate bar, etc.--, be creative! Avoid vases and microphones, or adding too many details to the object's description in general.
\end{lstlisting}
\end{tcolorbox}
\caption{Prompts used for generating a story context, after infilling the variables (number of people, containers, topics, names, and location). Names and location are sampled independently to increase diversity, prompts shown in Fig.~\ref{fig:prompt_names}.}
    \label{fig:prompt_story_setting}
\end{figure}

\begin{figure}[h]
    \centering
\begin{tcolorbox}[
        colback=promptboxcolor,    
        colframe=black,    
        boxrule=0.5pt,        
        arc=0mm,              
        left=1mm,             
        right=1mm,            
        top=1mm,              
        bottom=1mm]           
\lstset{
  basicstyle=\ttfamily,
  breaklines=true,
  breakatwhitespace=false,
  columns=fullflexible,
  keepspaces=true,
  frame=none,
}
\begin{lstlisting}[breakindent=0pt]
List 100 names. Do not include any other text.
\end{lstlisting}
\end{tcolorbox}
\begin{tcolorbox}[
        colback=promptboxcolor,    
        colframe=black,    
        boxrule=0.5pt,        
        arc=0mm,              
        left=1mm,             
        right=1mm,            
        top=1mm,              
        bottom=1mm]           
\lstset{
  basicstyle=\ttfamily,
  breaklines=true,
  breakatwhitespace=false,
  columns=fullflexible,
  keepspaces=true,
  frame=none,
}
\begin{lstlisting}[breakindent=0pt]
Suggest 100 different general contexts in which a story may happen. The context should be able to have several people in the same location easily listening and observing each other.

1. a school
2. a hospital
3. a vet shop
4. a family living room

Follow the format and make the descriptions as short as possible. Do not include any text before the list.
\end{lstlisting}
\end{tcolorbox}
\caption{Prompts used for generating a list of possible characters' names and locations for the story.}
\label{fig:prompt_names}
\end{figure}

\newpage
\subsection{Prompts used for story infilling}

\begin{figure}[h!]
    \centering
\begin{tcolorbox}[
        colback=promptboxcolor,    
        colframe=black,    
        boxrule=0.5pt,        
        arc=0mm,              
        left=1mm,             
        right=1mm,            
        top=1mm,              
        bottom=1mm]           
\lstset{
  basicstyle=\ttfamily,
  breaklines=true,
  breakatwhitespace=false,
  columns=fullflexible,
  keepspaces=true,
  frame=none,
}
\begin{lstlisting}[breakindent=0pt, literate={é}{{\'e}}1]
You are an expert writer that uses simple language, avoiding sounding unnatural or cliché. You are clear, creative, and helpful. You use simple sentence constructions and words so that everyone may understand you.
\end{lstlisting}
\end{tcolorbox}
\caption{System prompt used for story infilling}
    \label{fig:prompt_system}
\end{figure}
\begin{figure}[h!]
    \centering
\begin{tcolorbox}[
        colback=promptboxcolor,    
        colframe=black,    
        boxrule=0.5pt,        
        arc=0mm,              
        left=1mm,             
        right=1mm,            
        top=1mm,              
        bottom=1mm]           
\lstset{
  basicstyle=\ttfamily,
  breaklines=true,
  breakatwhitespace=false,
  columns=fullflexible,
  keepspaces=true,
  frame=none,
}
\begin{lstlisting}[breakindent=0pt]
Given the following story and knowing the description of the characters involved, write the start of a story. Don't actually describe any actions in the story, just the setting in which the story will happen. Only include the characters that are mentioned in the story.

STORY:
{story_script}

CHARACTERS:
{characters_description}

TWO-SENTENCE STORY BEGINNING THAT DOES NOT INCLUDE OR SUGGEST ANY INFORMATION OF WHAT WILL HAPPEN IN THE STORY. DO NOT MENTION PEOPLE:
\end{lstlisting} \end{tcolorbox}

\caption{System prompt used for sampling narration (the start of the story, before infilling).}
    \label{fig:prompt_context}
\end{figure}

\begin{figure}[h!]
    \centering
\begin{tcolorbox}[
        colback=promptboxcolor,    
        colframe=black,    
        boxrule=0.5pt,        
        arc=0mm,              
        left=1mm,             
        right=1mm,            
        top=1mm,              
        bottom=1mm]           
\lstset{
  basicstyle=\ttfamily,
  breaklines=true,
  breakatwhitespace=false,
  columns=fullflexible,
  keepspaces=true,
  frame=none,
}
\begin{lstlisting}[breakindent=0pt]
Given the following story and knowing the description of the characters involved, suggest a reasonable goal for each character. Only include the characters that were mentioned in the story.

STORY:
{story_script}

CHARACTERS:
{characters_description}

CHARACTERS GOALS: <insert here>

Follow the format and do not include any other text. Only include the characters mentioned in the story, and do not even mention the others in your list.
\end{lstlisting} \end{tcolorbox}
\caption{System prompt used for sampling character goals.}
    \label{fig:prompt_goals}
\end{figure}

\begin{figure}[h!]
    \centering
\begin{tcolorbox}[
        colback=promptboxcolor,    
        colframe=black,    
        boxrule=0.5pt,        
        arc=0mm,              
        left=1mm,             
        right=1mm,            
        top=1mm,              
        bottom=1mm]           
\lstset{
  basicstyle=\ttfamily,
  breaklines=true,
  breakatwhitespace=false,
  columns=fullflexible,
  keepspaces=true,
  frame=none,
}
\begin{lstlisting}[breakindent=0pt]
Continue the story {story_length}, clearly conveying the action or information below without altering it. Do not contradict any prior information. Avoid repeating the information verbatim, instead naturally (and possibly implicitly, but still unambiguously) conveying the meaning. Do not add characters or actions that were not explicitly described. Do not replace characters even if this would improve flow. Combining actions into a single sentence is OK as long as you do not alter the original information. {infilling_text_type}

Make it a short, yet an interesting story to read. Make the text exciting to read as well as each character's speech, so try to avoid e.g. starting all the sentences the same way. The story needs to follow common sense, e.g. do not magically change an object's location without mentioning it. Do not include any notes, comments, parentheses, or any other form of extra text that would not belong in a story. Feel free to hint or describe characters' goals and motivations for performing the actions if it would make the story flow better.

As a warning, take into account that when someone tells someone privately they might not be in the same location, e.g. they might be sending a text message or making a phone call; they might also be in the same location, in that case they could also communicate through a gesture, a whisper, etc. Do not assume a person is in the same room if it has not been made explicit before. Also, if someone was spying, or if they were distracted and did not listen or saw something happen, do not forget to include it! Remember that in this case, it should be clear that the distraction or spying applies only to the action mentioned and they go back to normal after the action is finished. Avoid making multiple copies of the same object.

Give {num_tries_completions} responses, ensuring to give {num_tries_completions} different phrasings of continuing the story conveying the action. Use very different wordings and sentence structures, but avoid changing object or room names!

WHO ARE THE CHARACTERS: {people_with_personas}

WHAT ARE THEIR GOALS: {optional_characters_goals}

NEW ACTION OR INFORMATION TO INCLUDE: {new_information}

CURRENT SHORT STORY: {story_context}

Follow the format and do not include any other text. Do not include any text before the list. Do not enumerate. Continue the story {story_length}. Avoid repeating the information verbatim, instead naturally (and possibly implicitly, but still unambiguously) conveying the meaning.

STORY CONTINUATION: <fill>

STORY CONTINUATION: <fill>
\end{lstlisting} \end{tcolorbox}
\caption{Prompt used for iterative story infilling including characters' goals, and allowing for simultaneous sampling of several possible infillings, which when associated with repetition penalty, yields more diverse infillings. Infilling length is uniformly chosen between \textit{`with a single sentence'} and \textit{`with up to two sentences'}, and infilling text type is uniformly chosen between \textit{`Make the new text be declarative, without including conversations.'} and \textit{`Make the new text conversational, using direct quotes to convey the words spoken by a character.'}}
    \label{fig:infilling_prompt}
\end{figure}


\newpage~\newpage

\section{\methodname{} examples  (cont. from \S\ref{eval_benchmark})}\label{app:generation_examples}

See a large sample of \methodname{}-generated data in \url{https://huggingface.co/datasets/facebook/ExploreToM}. We also include a few examples below.

\subsection{Story structure examples}

\begin{figure}[h!]
\begin{center}
{
\small
\centering
\noindent\fbox{%
    \parbox{0.95\linewidth}{%
\begin{itemize}[noitemsep,nolistsep,leftmargin=*]
\item Addison entered the monastery dining hall.
\item Addison filled the large ceramic vase with fresh sunflowers.
\item Addison left the monastery dining hall.
\item Charlotte entered the monastery dining hall.
\item Charlotte painted the large ceramic vase with intricate designs in gold.
\item Charlotte glued a few loose diamonds around the neck of the large ceramic vase. While this action was happening, Addison witnessed this action in secret (and only this action).
\end{itemize}
    }%
}
}
\end{center}
\begin{center}
{
\small
\centering
\noindent\fbox{%
    \parbox{0.95\linewidth}{%
\begin{itemize}[noitemsep,nolistsep,leftmargin=*]
\item Amelia entered the staff room.
\item Amelia moved the large first aid kit to the plastic storage bin, which is also located in the staff room. While this action was happening, Alexis witnessed this action in secret (and only this action).
\item Amelia entered the equipment storage room. 
\item Amelia left the equipment storage room. 
\item Amelia entered the staff room. 
\item Amelia moved the large first aid kit to the equipment storage room, leaving the plastic storage bin in its original location.
\item Amelia moved the large first aid kit to the metal cabinet, which is also located in the equipment storage room.
\end{itemize}
    }%
}
}
\end{center}
\begin{center}
{
\small
\centering
\noindent\fbox{%
    \parbox{0.95\linewidth}{%
\begin{itemize}[noitemsep,nolistsep,leftmargin=*]
\item Alexander entered the city hall planning department. 
\item Alexander moved the large map of the city to the cardboard tube, which is also located in the city hall planning department. 
\item Alexander told privately to Leslie that the large map of the city is in the cardboard tube. 
\item Victoria entered the city hall planning department. 
Leslie told privately to Peyton that the large map of the city is in the cardboard tube. 
\item Victoria moved the large map of the city to the plastic storage bin, which is also located in the city hall planning department.
\end{itemize}
    }%
}
}
\end{center}
\caption{Examples of the following story structures: $a_{\text{updateObjState}}$ ($\mathcal{A}_2$) with asymmetry; $\{a_{\text{moveObjContainer}}, a_{\text{moveObjRoom}}\}$ ($\mathcal{A}_4$) with asymmetry; and $\{a_{\text{moveObjContainer}, a_{\text{info}}}\}$  ($\mathcal{A}_5$) without asymmetry.}
\end{figure}

\newpage
\subsection{Story infilling examples}
\begin{table}[h]
\caption{Example of adversarially sampled story structure and iterative infilling with Llama-3.1 70B Instruct, including a story context (first row). This example showcases how the infilling process is able to group sentences when they depict the same event but for more than one person, allowing for an even more natural-sounding story.}
\centering
\begin{tabular}{@{}p{2in}p{3.2in}@{}}
\toprule
Story Structure (list of actions)                                                                     & Infilling for each action                                                           \\ \toprule \midrule
$<$initial context$>$ & The operations center, a bustling hub of activity during the park's peak festival season, was dimly lit except for a few overhead fluorescent lights that cast a sterile glow over the rows of workstations and storage units. The air was thick with the smell of stale coffee and fresh printer paper, a familiar scent to anyone who spent time behind the scenes of the busy amusement park. \\ \midrule
Brooklyn entered the operations center. & \multirow{2}{3.2in}{As she entered the operations center, Brooklyn's presence was swiftly followed by James, whose rugged demeanor stood in contrast to the sterile environment of the room.} \\ \cdashline{1-1} 
James entered the operations center. &  \\ \midrule
James moved the large festival schedule poster to the cardboard box, which is also located in the operations center. &  James neatly relocated the oversized festival schedule poster, carefully placing it within the confines of a cardboard box, a receptacle that had been securely positioned in a corner of the operations center. \\ \midrule
James left the operations center. & \multirow{2}{3.2in}{Brooklyn stepped out into the bright sunlight, a stark contrast to the dimly lit operations center she had just left, while James followed suit, the warm rays illuminating his weathered features. The festival preparations were now unfolding across the park, with both of them poised to tackle their respective tasks.} \\ \cdashline{1-1}
\begin{tabular}[c]{@{}l@{}}Brooklyn left the operations center. \\ \\  \\ \\ \\ \end{tabular}  &     \\ \midrule
Tyler entered the operations center. & \multirow{2}{3.2in}{As Tyler walked into the operations center, the soft hum of idle machinery and the faint scent of printer paper enveloped him, a familiar yet calming atmosphere before the chaos of the festival. The room's sterile lighting seemed to intensify, reflecting off the polished surfaces as Brooklyn slipped in, her footsteps quiet on the worn floor.} \\ \cdashline{1-1}
\begin{tabular}[c]{@{}l@{}}Brooklyn entered the operations \\  center. \\ \\ \\  \\ \end{tabular}  &     \\ \midrule
Brooklyn moved the large festival schedule poster to the plastic storage bin, which is also located in the operations center. & Brooklyn's hands moved swiftly as she carefully slid the oversized festival schedule poster into the large plastic storage bin in the operations center.
 \\ \bottomrule
\end{tabular}

\end{table}

\begin{table}[h]
\caption{Example of adversarially sampled story structure and simple iterative infilling with Llama-3.1 70B Instruct, along with example questions and model answers.}
\centering
\begin{tabular}{@{}p{2in}p{3.2in}@{}}
\toprule
Story Structure (list of actions)                                                                     & Infilling for each action                                                                                                                                                                                       \\ \toprule \midrule
David entered the study room. & \multirow{2}{3.2in}{David walked into the study room, scanning the organized chaos of files on the desk, just as Sarah stepped in, clutching her tablet and a determined look.   } \\ \cdashline{1-1}
\begin{tabular}[c]{@{}l@{}}Sarah entered the study room. \\ \\  \end{tabular}  &     \\ \midrule
Sarah moved the prototype model to the metal filing cabinet, which is also located in the study room. & Sarah moved the prototype model to the metal filing cabinet, her fingers brushing against the cool surface before turning to David with a confident smile.                                                      \\ \midrule
David left the study room.                                                                            & After a brief exchange with Sarah, David grabbed his briefcase and left the study room, mindful of the upcoming meeting downtown.                                                                               \\ \midrule
Mark entered the study room.                                                                          & Mark entered the study room, his eyes quickly assessing the scattered documents and half-finished sketches on the table, before settling into his favorite chair by the window to brainstorm new ideas.         \\ \midrule
Mark moved the prototype model to the wooden chest, which is also located in the study room.          & Mark, contemplating his latest concept, carefully picked up the prototype model from the metal filing cabinet and transferred it to the wooden chest, a better place for such an inspiration-spurring creation. \\ \bottomrule
\end{tabular}

\vspace{2pt}
\centering
\begin{tabular}{@{}p{1.8in}p{1.1in}p{1.5in}p{0.5in}@{}}
\toprule
Selected Questions            & \begin{tabular}[c]{@{}l@{}}\methodname{}\\-generated answer\\ using our domain-\\specific language\end{tabular} & GPT-4o's responses   & Correct?    \\ \toprule
In which container will David search for the prototype model? & metal filing cabinet                                                                & david will search for the prototype model in the metal filing cabinet. & Yes       \\
In which container will Sarah search for the prototype model? & wooden chest                                                                        & sarah will search for the prototype model in the metal filing cabinet. & No         \\
In which container will Mark search for the prototype model?  & wooden chest                                                                        & mark will search for the prototype model in the wooden chest.          & Yes       \\
In which container will David search for the prototype model? & metal filing cabinet                                                                & the metal filing cabinet.                                              & Yes        \\
In which container will Sarah search for the prototype model? & wooden chest                                                                        & sarah will search for the prototype model in the metal filing cabinet. & No    \\
In which container will Mark search for the prototype model?  & wooden chest                                                                        & the wooden chest.                                                      & Yes                \\ \bottomrule
\end{tabular}

\end{table}

\end{document}